\documentclass{article} 
\usepackage{iclr2027_conference,times}

\usepackage{amsmath,amsfonts,bm}

\def\eqref#1{equation~\ref{#1}}

\def\1{\bm{1}}

\DeclareMathAlphabet{\mathsfit}{\encodingdefault}{\sfdefault}{m}{sl}
\SetMathAlphabet{\mathsfit}{bold}{\encodingdefault}{\sfdefault}{bx}{n}

\usepackage{hyperref}
\usepackage{url}
\DeclareUnicodeCharacter{266B}{\mbox{\usefont{U}{wasy}{m}{n}\char15}}
\usepackage{graphicx}
\usepackage{float}
\usepackage{booktabs}
\title{ARC-KV: Amortizing Anchor Search for Reconstruction-Based KV Cache Compaction}

\author{%
Zheyu Shen$^{1}$\thanks{This work was done during an internship at Meta.}\quad
Guanhua Wang$^{2}$\thanks{Corresponding author: \texttt{guanhua@meta.com}.}\quad
Dezhan Tu$^{2}$\quad Mengchi Zhang$^{2}$\\
\bfseries Yanjia Li$^{2}$\quad Adnan Aziz$^{2}$\quad Chunqiang Tang$^{2}$\quad Ang Li$^{1}$\\
\normalfont $^{1}$University of Maryland, College Park (\href{https://umd.edu}{umd.edu})\\
\normalfont $^{2}$Meta (\href{https://meta.com}{meta.com})\\
\normalfont\texttt{\{zyshen,angliece\}@umd.edu}\\
\normalfont\texttt{\{guanhua,dztu,mengchi,ethanyanjiali,adnanaziz,tang\}@meta.com}%
}

\iclrfinalcopy 
\begin{document}

\maketitle
\ificlrfinal\lhead{Preprint}\fi

\begin{abstract}
Long-context large language model inference is bottlenecked by KV caches that grow linearly with sequence length.
This burden is especially severe for long, reusable context prefixes, whose cache must serve many downstream queries.
Reconstruction-based methods such as Attention Matching achieve strong downstream task performance with compact KV caches.
However, iterative anchor search dominates the compaction cost of OMP-based Attention Matching.
This motivates our selective amortization principle of learning a reusable anchor-selection policy across contexts while retaining context-specific reconstruction.
In this work, we propose ARC-KV, a novel reconstruction-based KV cache compaction method that follows this principle. To this end, we first train a value-aware indexer to select real-key anchors in a single scoring pass. ARC-KV then applies convex-hull-constrained key merging and fits an attention-mass bias and compact values against the full cache. At inference time, ARC-KV builds the compact cache once per context using the frozen indexer and reuses it for all subsequent queries.
Extensive experiments demonstrate that ARC-KV outperforms reported compaction methods in most settings across QuALITY, RULER, and LongBench on Llama-3.1-8B-Instruct. In particular, at 10\% KV retention on QuALITY, ARC-KV improves accuracy from 0.6409 to 0.6474 over Attention Matching while reducing compaction time by a factor of 25.73, from 959.8\,s to 37.3\,s.


\end{abstract}

\section{Introduction}
The rapid evolution of large language models (LLMs) has unlocked remarkable capabilities across a wide array of tasks, including reasoning, coding and agentic workflows ~\citep{openai2026gpt5, yangQwen3Technical2025, guo2025deepseek}. Expanding the context window beyond 1M tokens has further extended these capabilities. For instance, Gemini Pro ~\citep{comanici2025gemini} supports a 10-million-token context window, enabling deep debugging over entire system logs, stack traces, and source code.

However, these gains come at the cost of intensive computation and memory consumption, primarily stemming from the attention mechanism. Since the Key-Value (KV) cache grows significantly as sequence length increases, the auto-regressive decoding stage must repeatedly write and read this large KV cache to and from memory, making the process memory-bound and posing a major challenge to scaling long-context LLMs. Take LLaMA-65B as an example: a 128K-token sequence requires 320GB of memory for the KV cache alone in FP16~\citep{hooperKVQuantTowards2024}.
Such a memory footprint of the KV cache degrades not only serving throughput but also latency.

Extensive prior work has leveraged attention sparsity to accelerate long-context inference. These approaches fall into three main categories: \textit{rule-based token eviction}, \textit{learned token eviction}, and \textit{cache reconstruction}. \textit{Rule-based token eviction} method discards KV pairs according to token position or importance score, such as StreamingLLM~\citep{xiaoStreamingLLM2024}, H2O~\citep{zhangH2OHeavyHitterOracle2023} and SnapKV~\citep{liSnapKVLLMKnows2024}; \textit{Learned eviction} policies (e.g., KVP~\citep{moschellaLearningEvictKeyValue2026}, LKV~\citep{zhouLKVEndtoEndLearning2026}) train a lightweight selector to retain a subset of the KV cache. However, these policies permanently evict tokens, rendering them inaccessible in all subsequent decoding steps and causing irreversible information loss.


The most recent \textit{cache reconstruction} approach addresses the limitations of direct eviction policies by reconstructing the KV cache into a compact form that preserves rich information. This line of work includes Cartridges~\citep{eyubogluCartridgesLightweightGeneralpurpose2025}, Attention Matching (AM)~\citep{zweigerFastKVCompaction2026}, and Still~\citep{oneillStillAmortizedKV2026}. Notably, AM consistently outperforms rule-based token eviction policies even at high compaction ratios (20×–100×). Its OMP-based variant uses orthogonal matching pursuit (OMP)~\citep{patiOrthogonalMatchingPursuit1993} to select original keys through iterative search and attention-mass refitting, then fits compact values to match the full cache's attention output. Still instead trains a compactor to synthesize the compact cache in one forward pass.


OMP-based AM makes a coupled design choice by treating both anchor discovery and cache reconstruction as per-context optimization problems.
Our key insight is that these operations have different generalization requirements. A selection policy can learn a reusable prior for identifying effective anchors across contexts of the same frozen model. Reconstruction must approximate the attention mass and outputs of the current cache, motivating direct fitting to that context. We therefore amortize the selection policy across contexts while retaining context-specific reconstruction.

We instantiate this decomposition in ARC-KV. First, \textit{Anchor Selection} utilizes a value-aware indexer to condition on reference activations, queries, and cached KV pairs, scoring and selecting anchors in a single pass. Second, \textit{Deterministic Merging} groups keys based on their responses to reference queries and shifts each anchor toward its group's mass-weighted centroid, absorbing non-anchor information while constraining each compact key to its convex hull. Third, \textit{Closed-Form Fitting} leverages context-specific equations to solve for \(\beta \) and \(C_{v}\) using fixed compact keys. Our contributions are summarized as follows:





\begin{itemize}
    \item We identify anchor search as a component of reconstruction-based KV compaction that can be amortized across contexts through a reusable selection policy while keeping reconstruction context-specific. Based on this decomposition, we introduce ARC-KV, a selectively amortized KV-cache compaction framework for reusable context prefixes.
    \item We instantiate this framework with a value-aware indexer whose reconstruction-training objective targets held-out post-fitting attention-output fidelity using disjoint scoring, fitting, and holdout queries, without OMP-generated anchor labels. At inference, response-profile key merging and analytical attention-mass and compact-value fitting construct the compact cache from the selected anchors.
    \item We evaluate the complete ARC-KV pipeline on QuALITY, RULER, and LongBench v1 with Llama-3.1-8B-Instruct, where it ranks first among the reported compacted methods in most evaluated settings. On QuALITY at 10\% nominal retention, it reduces reported compaction time by a factor of 25.73 relative to OMP-based AM while increasing accuracy from 0.6409 to 0.6474.
\end{itemize}

\section{Motivation and Analysis}
\label{sec:background}

In this section, we empirically analyze cache reconstruction fidelity, iterative anchor search efficiency and value-influence token ranking. We reveal that (1) a hard KV subset cannot retain sufficient information, (2) iterative anchor search dominates the computational cost, and (3) value vectors provide a consistent signal beyond attention-only anchor scores.

\subsection{Cache Reconstruction Outperforms Token Eviction in Fidelity}
Token eviction selects an index set $S$ and retains the KV pairs $(K_S,V_S)$ unchanged, so a query $q$ attends through $\operatorname{softmax}(qK_S^\top)V_S$; reconstruction-based methods instead use the selected keys as anchors to build compact keys $C_k$, a mass bias $\beta$, and compact values $C_v$ and evaluate $\operatorname{softmax}(qC_k^\top+\beta^\top)C_v$ (Section~\ref{sec:problem_formulation}; Appendix~\ref{app:background_operators}).

To isolate representation capacity from anchor quality and ensure a fair comparison, we fix the same 5\% anchor set across all five cache constructions. As shown in Table~\ref{tab:motivation_reconstruction}, the comparison then separately tests mass calibration, key or value merging, and compact value fitting, all while leaving the anchor indices unchanged.

\begin{table}[t]
\caption{Held-out-query reconstruction scores (mean $\pm$ SD over 2,560 article--layer--KV-head cells) on QuALITY with Llama-3.1-8B-Instruct at 5\% retention.
All constructions use the same fixed anchors.
The key/value-merging row is diagnostic; ARC-KV merges only keys and fits $C_v$ globally.}
\label{tab:motivation_reconstruction}
\centering
\footnotesize
\setlength{\tabcolsep}{3.5pt}
\begin{tabular}{@{}l|c|c|c|c|c@{}}
\toprule
Construction & $C_k$ & $\beta$ & $C_v$ & Rel. $\ell_2$ $\downarrow$ & Cosine $\uparrow$ \\
\midrule
Hard subset & $K_S$ & $0$ & $V_S$ & $0.8778{\pm}0.2973$ & $0.7543{\pm}0.1055$ \\
Mass calibration & $K_S$ & fitted & $V_S$ & $0.8782{\pm}0.3189$ & $0.7503{\pm}0.1100$ \\
Key/value merging & $C_k^{\mathrm{mrg}}$ & fitted & $C_v^{\mathrm{mrg}}$ & $0.5109{\pm}0.1481$ & $0.8558{\pm}0.0738$ \\
Value fitting & $K_S$ & fitted & $C_v^{\mathrm{LS}}$ & $0.3546{\pm}0.1265$ & $0.9135{\pm}0.0586$ \\
Key merging $+$ value fitting & $C_k^{\mathrm{mrg}}$ & fitted & $C_v^{\mathrm{LS}}$ & $\mathbf{0.3200{\pm}0.1291}$ & $\mathbf{0.9250{\pm}0.0566}$ \\
\bottomrule
\end{tabular}
\end{table}

\noindent\begin{minipage}{\textwidth}
The hard subset yields a relative $\ell_2$ error of $0.8778$, and fitting $\beta$ alone leaves the local output error essentially unchanged.
Fitting $C_v$ reduces the error to $0.3546$, showing that the original anchor values impose a substantial representation constraint in this setting.
Merging keys before fitting $C_v$ gives the lowest error of $0.3200$.
Thus, selected keys should serve as anchors for constructing the compact cache $(C_k,\beta,C_v)$, rather than be retained unchanged as the final compact cache.
\end{minipage}

\subsection{Iterative Anchor Search Dominates Compaction Time}
We profile the official AM implementation~\citep{zweigerFastKVCompaction2026}, split into reference-query construction, OMP~\citep{patiOrthogonalMatchingPursuit1993} anchor search, $\beta$ fitting, and $C_v$ fitting (Appendix~\ref{app:am_profile}, Figure~\ref{fig:motivation_results}b).
At 20\% retention, AM's accelerated OMP search (OMP-fast) alone needs 3.70 minutes at 4K tokens and 15.36 hours at 64K, while every other stage takes at most 1.17 minutes at 64K; amortization should therefore target anchor search.

\subsection{Value Vectors Matter in Token Selection}
Attention-only selectors (HA)~\citep{zhangH2OHeavyHitterOracle2023,liSnapKVLLMKnows2024} rank keys by attention probabilities alone, whereas value-influence ranking (VI)~\citep{guoVATP2024,goelCAOTE2025} uses each key's normalized leave-one-out effect on the attention output.
Under a matched budget and disjoint held-out queries, VI lowers truncated-output error at every keep ratio, by 1.05--6.23\% relative to HA on four layer--KV-head pairs (Appendix~\ref{app:value_selector_details}, Figure~\ref{fig:motivation_results}a).


\section{Method}
\label{sec:method}

ARC-KV replaces iterative, per-context anchor search with a value-aware indexer that is trained once per base model and then frozen; for each new context it selects real-key anchors in one scoring pass, merges discarded keys into them, and analytically fits an attention-mass bias and compact values.
We describe one layer--KV-head pair under grouped-query attention (GQA); the same procedure is applied independently to all pairs.

\subsection{Problem Formulation}
\label{sec:problem_formulation}

Let $K,V\in\mathbb{R}^{T\times d}$ denote the key and value cache for a context of length $T$, with rotary position embedding (RoPE)~\citep{suRoFormerEnhancedTransformer2024} applied to the keys.
We write $k_j$ and $v_j$ for their $j$-th rows and $[T]=\{1,\ldots,T\}$ for the cache positions.
Throughout the paper, we omit the usual inverse-square-root scaling from attention-softmax logits.
For queries $Q\in\mathbb{R}^{n\times d}$, the full-cache attention probabilities and outputs are $P(Q;K)=\operatorname{softmax}(QK^\top)$ and $O(Q;K,V)=P(Q;K)V$.
Given a budget $t\ll T$ and keep ratio $\rho=t/T$, we seek an anchor set $S=\{s_r\}_{r=1}^{t}\subset[T]$ and a compact representation with $t$ entries.
It consists of compact keys $C_k=[c_{k,1};\ldots;c_{k,t}]\in\mathbb{R}^{t\times d}$, an attention-mass bias $\beta\in\mathbb{R}^{t}$, and compact values $C_v\in\mathbb{R}^{t\times d}$.
Each selected key $k_{s_r}$ serves as a real-key anchor that defines a group of cache positions.
The corresponding compact key $c_{k,r}$ aggregates key-side information within that group and is constrained to its convex hull, while $\beta$ and $C_v$ are fitted against the full cache.
Thus, unlike hard retention, ARC-KV need not set $C_k=K_S$ or $C_v=V_S$, where $K_S$ and $V_S$ collect the rows of $K$ and $V$ indexed by $S$.
The resulting compact operator is
\begin{equation}
\widehat O(Q;C_k,\beta,C_v)
=\operatorname{softmax}\!\left(
QC_k^\top+\mathbf{1}_n\beta^\top
\right)C_v.
\label{eq:compact_attention}
\end{equation}

Preserving only the compacted context's locally normalized attention output is insufficient once it is attended together with uncompressed KV states.
For a query $q$, define the unnormalized attention mass of the original and compact caches as
\begin{equation}
Z_K(q)=\sum_{j=1}^{T}\exp(qk_j^\top),
\qquad
Z_C(q;\beta)=\sum_{r=1}^{t}\exp(qc_{k,r}^\top+\beta_r).
\label{eq:block_attention_mass}
\end{equation}
Fitting $\beta$ so that $Z_C(q;\beta)\approx Z_K(q)$ preserves how strongly the compacted context competes with tokens outside the compacted block, while fitting $C_v$ preserves its attention outputs.
Appendix~\ref{app:joint_objective} states the joint objective when the compact block is attended together with uncompressed KV states.
Our goal is therefore to train the value-aware indexer to select anchors from which $(C_k,\beta,C_v)$ can be constructed to preserve the full cache's attention mass and attention outputs for unseen inference queries.

\subsection{Method Overview}
\label{sec:method_overview}

ARC-KV constructs a compact cache in four context-specific stages.
First, a frozen language model produces the full cache $(K,V)$ and a pool of reference activation--query pairs.
Second, the trained value-aware indexer scores every cache position, and top-$t$ selection produces the real-key anchor set $S$.
Third, every cached key is assigned to the anchor with the most similar normalized attention-response profile, and each group is merged into one compact key.
Fourth, with the merged $C_k$ fixed, analytical fitting determines the attention-mass bias $\beta$ and compact values $C_v$.

\begin{figure}[t]
\centering
\includegraphics[width=\linewidth]{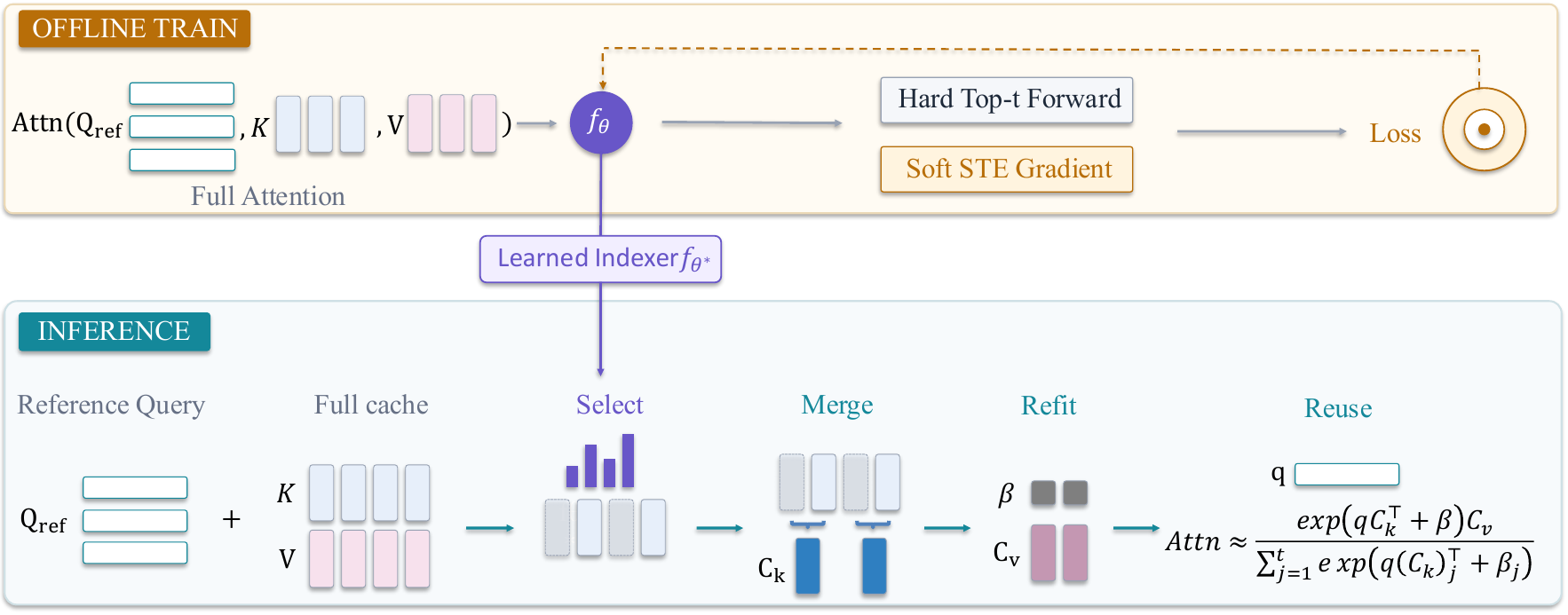}
\caption{Overview of ARC-KV. Full-attention supervision trains a reusable indexer with hard top-$t$ selection and soft surrogate gradients. At inference, selected anchors undergo context-specific key merging and fitting of $\beta$ and $C_v$.}
\label{fig:method_overview}
\end{figure}

Only the indexer parameters are learned (Section~\ref{sec:indexer_training}); the same frozen indexer is reused across contexts, while reference-query construction, scoring, merging, and fitting are performed for each context (Figure~\ref{fig:method_overview}).

\paragraph{Reference-query pool.}
Following KVzip~\citep{kimKVzipQueryAgnosticKV2025} and Attention Matching~\citep{zweigerFastKVCompaction2026}, reference queries come from repeat-prefill: after the initial prefill the model is asked to repeat the context, and we retain each repeated position's pre-query-projection activation $x_i$ and post-RoPE query $q_i$, forming $\mathcal{D}_{\mathrm{ref}}=\{(x_i,q_i)\}_{i=1}^{n_{\mathrm{ref}}}$ with row-stacked $X_{\mathrm{ref}}$ and $Q_{\mathrm{ref}}$ (Appendix~\ref{app:reference_queries}).
All compacted keys and values come from the original-context prefill.

\subsection{Value-Aware Indexer}
\label{sec:key_indexer}

The value-aware indexer is a lightweight scoring model whose role is to replace iterative anchor search with one scoring pass.
For each reference pair $(x_i,q_i)$ and full cache $(K,V)$, it produces one logit for every cache position,
$I_\theta(x_i,q_i,K,V)=[I_{i,1},\ldots,I_{i,T}]\in\mathbb{R}^{T}$.
Under GQA, each KV head is shared by multiple query heads. When scoring a KV head, we use the reference queries from all query heads mapped to it.

We extend the low-dimensional query--key scorer of the DeepSeek Sparse Attention indexer~\citep{deepseek-aiDeepSeekV32PushingFrontier2025} with one auxiliary value-attention head and represent each position by both $k_j$ and $v_j$, so keys with similar attention patterns can score differently when their values differ (architecture in Appendix~\ref{app:indexer_architecture}).

To obtain one context-level ranking, we apply the indexer to $n_{\mathrm{score}}$ reference pairs.
For each row $i$, we normalize the indexer logits $I_{i,:}$ across the $T$ cache positions.
We then aggregate each position across rows by root-mean-square~\citep{rehgKVCompress2024,zweigerFastKVCompaction2026}:
\begin{equation}
\pi_{i,j}=\frac{\exp(I_{i,j})}{\sum_{u=1}^{T}\exp(I_{i,u})},
\qquad
s_j=\left(\frac{1}{n_{\mathrm{score}}}\sum_{i=1}^{n_{\mathrm{score}}}\pi_{i,j}^2\right)^{1/2},
\label{eq:indexer_pooling}
\end{equation}
Root-mean-square pooling gives additional weight to large query-specific probabilities, retaining positions that are important to only a subset of reference queries.
We select the $t$ positions with the largest aggregate scores as real-key anchors, $S=\operatorname{TopK}_{j\in[T]}(s_j,t)$.
Unlike query-conditioned sparse attention, ARC-KV selects this set once per context and layer--KV-head pair and reuses it for all future queries.

\subsection{Convex-Hull-Constrained Key Merging}
\label{sec:key_merging}

Selecting anchors alone discards the key-side geometry of the remaining $T-t$ positions.
We recover part of this information with a deterministic post-selection merge that groups keys by their normalized attention-response profiles.
Unlike distance in the raw key space, profile similarity directly compares how keys respond to the same reference queries.
A schematic is given in Appendix~\ref{app:key_merging_details} (Figure~\ref{fig:key_merging_schematic}).

We use all reference queries $Q_{\mathrm{merge}}=Q_{\mathrm{ref}}=[q_i]_{i=1}^{n_{\mathrm{ref}}}$ and define each normalized attention-response profile $u_j$ from the post-RoPE attention logits:
\begin{equation}
\ell_{i,j}=q_i^\top k_j,
\qquad
z_{\mathrm{merge}}=\max_{i,j}\ell_{i,j},
\qquad
u_j=\frac{[\exp(\ell_{i,j}-z_{\mathrm{merge}})]_{i=1}^{n_{\mathrm{ref}}}}
{\lVert[\exp(\ell_{i,j}-z_{\mathrm{merge}})]_{i=1}^{n_{\mathrm{ref}}}\rVert_2}.
\label{eq:response_profile}
\end{equation}
Each key is hard-assigned to the anchor with the most similar normalized attention-response profile:
\begin{equation}
\kappa(s_r)=r,
\qquad
\kappa(j)=\operatorname*{arg\,max}_{r\in[t]}u_j^\top u_{s_r}
\quad (j\notin S),
\qquad
G_r=\{j\in[T]:\kappa(j)=r\}.
\label{eq:anchor_assignment}
\end{equation}
Every group therefore contains its own anchor and is nonempty.

Within each group, we define the key-merge weights from query-averaged unnormalized attention mass:
\begin{equation}
\phi_j=\frac{1}{n_{\mathrm{ref}}}\sum_{i=1}^{n_{\mathrm{ref}}}\exp(\ell_{i,j}-z_{\mathrm{merge}}),
\qquad
\mu_r=\frac{\sum_{j\in G_r}\phi_jk_j}{\sum_{j\in G_r}\phi_j},
\label{eq:mass_centroid}
\end{equation}
and move the anchor toward this weighted centroid:
\begin{equation}
c_{k,r}=(1-\lambda_m)k_{s_r}+\lambda_m\mu_r,
\qquad
0\leq\lambda_m\leq1.
\label{eq:key_merge}
\end{equation}
Here, $\lambda_m$ is the key-merging coefficient; $\lambda_m=0$ leaves $C_k=K_S$, and $\lambda_m=1$ replaces each anchor by its group centroid.
Because $s_r\in G_r$, $c_{k,r}\in\operatorname{conv}\{k_j:j\in G_r\}$, so for any query $q$ its key-derived logit lies between the minimum and maximum real-key logits in that group.
The convex-hull constraint therefore keeps each compact key within the geometry of its assigned real keys rather than extrapolating beyond them.
The merge operates only on $Q_{\mathrm{merge}}$ and $K$: values influence anchor selection through the value-aware indexer and enter the subsequent compact value fitting, but are not directly averaged within $G_r$.

\paragraph{Attention mass and compact value fitting.}
Given the merged $C_k$ and calibration queries $Q_{\mathrm{fit}}$ drawn from $\mathcal{D}_{\mathrm{ref}}$, we apply the closed-form calibration of Attention Matching~\citep{zweigerFastKVCompaction2026}: a projected least-squares solve finds nonnegative $w^\star$ with $\exp(Q_{\mathrm{fit}}C_k^\top)w^\star\approx\exp(Q_{\mathrm{fit}}K^\top)\mathbf{1}_T$ and sets $\beta^\star=\log w^\star$; a ridge regression then fits $C_v^\star$ (Appendix~\ref{sec:cache_reconstruction}).
Neither fit adds learned parameters.

\subsection{Training the Reusable Value-Aware Indexer}
\label{sec:indexer_training}

We train one indexer per frozen base model using the model's full-attention behavior as self-supervision, without anchor labels generated by iterative search.
Joint training optimizes reconstruction of held-out attention outputs through a hard-selection forward branch and a differentiable soft surrogate for backpropagation.
Both branches operate on real keys before key merging, so the learned objective targets anchor quality directly.

\paragraph{Reference-query episodes.}
For each training context, we sample three mutually disjoint subsets $\mathcal{D}_{\mathrm{score}}$, $\mathcal{D}_{\mathrm{fit}}$, and $\mathcal{D}_{\mathrm{holdout}}$ from $\mathcal{D}_{\mathrm{ref}}$ without replacement.
The scoring subset produces the context-level anchor ranking.
The fitting subset supplies $Q_{\mathrm{fit}}$ for attention mass and compact value fitting, while the holdout subset supplies $Q_{\mathrm{holdout}}$ for evaluating the reconstructed attention outputs.
This separation evaluates each selected and fitted cache on queries not used for either operation.
The three subsets are resampled deterministically at each epoch.

\paragraph{Reconstruction of held-out attention outputs.}
For a query $q$, let $p(q)=\operatorname{softmax}(qK^\top)$ and $y(q)=p(q)V$ denote the full-cache attention distribution and output.
The hard branch evaluates the discrete anchors used before inference-time key merging.
For a training cache of length $T$, we set $t=\max(1,\lceil\rho T\rceil)$, aggregate scores over $\mathcal{D}_{\mathrm{score}}$, and select $S=\operatorname{TopK}(s,t)$.
We set $C_k^{\mathrm{h}}=K_S$, then obtain $\beta^{\mathrm{h}}$ from Appendix Equation~\ref{eq:mass_fit} and $C_v^{\mathrm{h}}$ from Appendix Equation~\ref{eq:value_lstsq} using $Q_{\mathrm{fit}}$.
The reconstructed hard-branch output is $\widehat y^{\mathrm{h}}(q)=\widehat O(q;C_k^{\mathrm{h}},\beta^{\mathrm{h}},C_v^{\mathrm{h}})$.
With $n_{\mathrm{holdout}}=|\mathcal{D}_{\mathrm{holdout}}|$, its mean squared relative error is
\begin{equation}
\mathcal{L}_{\mathrm{out}}^{\mathrm{h}}
=\frac{1}{n_{\mathrm{holdout}}}
\sum_{i=1}^{n_{\mathrm{holdout}}}
\frac{\left\|\widehat y^{\mathrm{h}}(q_i)-y(q_i)\right\|_2^2}
{\left\|y(q_i)\right\|_2^2}.
\label{eq:output_loss}
\end{equation}
The complete hard branch is evaluated without gradient tracking.
It determines the forward loss value but contributes no gradient to the indexer.

Because top-$t$ selection is non-differentiable, the backward branch relaxes it over $\mathcal{C}=\operatorname{TopK}(s,\min\{T,2t\})$: each candidate receives a sigmoid gate $g_j$ on its score margin to the top-$t$ threshold (temperature annealed from $1.0$ to $0.1$), $\log g$ is added to the attention biases, $C_v^{\mathrm{s}}$ comes from a differentiable ridge solve on $Q_{\mathrm{fit}}$, and $\mathcal{L}_{\mathrm{out}}^{\mathrm{s}}$ is evaluated on $Q_{\mathrm{holdout}}$ as in Equation~\ref{eq:output_loss} (Appendix~\ref{app:indexer_training_details}).

We combine the hard forward value with the gradient of the soft surrogate through
\begin{equation}
\mathcal{L}_{\mathrm{out}}^{\mathrm{STE}}
=\operatorname{sg}\!\left(\mathcal{L}_{\mathrm{out}}^{\mathrm{h}}\right)
+\mathcal{L}_{\mathrm{out}}^{\mathrm{s}}
-\operatorname{sg}\!\left(\mathcal{L}_{\mathrm{out}}^{\mathrm{s}}\right).
\label{eq:output_ste}
\end{equation}
Its forward value is $\mathcal{L}_{\mathrm{out}}^{\mathrm{h}}$, while its gradient comes from $\mathcal{L}_{\mathrm{out}}^{\mathrm{s}}$ through the soft gates and differentiable compact-value fit.

\paragraph{KL warm-up and joint objective.}
Before joint training, we warm up the indexer by matching its query-specific key distribution to full attention on $\mathcal{D}_{\mathrm{ref}}$, minimizing $\mathcal{L}_{\mathrm{KL}}$, the mean over $(x,q)\in\mathcal{D}_{\mathrm{ref}}$ of $\operatorname{KL}\!\left(p(q)\,\|\,\operatorname{softmax}(I_\theta(x,q,K,V))\right)$ (Appendix Equation~\ref{eq:indexer_kl}).
The subsequent joint objective is
\begin{equation}
\mathcal{L}
=\lambda_{\mathrm{out}}\mathcal{L}_{\mathrm{out}}^{\mathrm{STE}}
+\lambda_{\mathrm{KL}}\mathcal{L}_{\mathrm{KL}}.
\label{eq:training_objective}
\end{equation}
The KL term provides dense query-specific attention supervision, while the reconstruction term supplies the value-aware signal by evaluating outputs $y(q)=p(q)V$ after analytical fitting.
The stage-specific weights $\lambda_{\mathrm{out}}$ and $\lambda_{\mathrm{KL}}$ are given in Section~\ref{sec:experimental_setup} and Appendix~\ref{app:training_details}.
Both objectives use full-attention behavior rather than anchors generated by OMP.
The base language model remains frozen, and only the indexer parameters $\theta$ are updated.

\paragraph{Inference.}
The pool is not partitioned at inference: the same reference pairs serve anchor scoring, key merging, and fitting.
The nominal budget is distributed over layer--KV-head pairs with fixed model-specific weights (Appendix~\ref{sec:method_inference}, Equation~\ref{eq:per_head_budget}), and long contexts select anchors within chunks (Appendix~\ref{app:kv_chunking}); $(C_k,\beta,C_v)$ is built once per context and reused for all future queries.
Our current scope is one-shot compaction of a reusable context prefix.

\section{Experiments and Results}

\subsection{Experimental Setup}
\label{sec:experimental_setup}

\paragraph{Model and implementation.}
We used Llama-3.1-8B-Instruct~\citep{grattafioriLlama3Herd2024} for the main evaluation and additionally assessed cross-model generalization with Qwen3-8B~\citep{yangQwen3Technical2025} and Gemma-3-12B-IT~\citep{gemmaTeamGemma3Technical2025} on QuALITY and RULER.
Because Gemma-3-12B-IT uses hybrid attention, we compacted only its eight full-attention layers (Appendix~\ref{app:setup_details}).
Its 40 sliding-window layers remain unchanged, so the retention ratio applies only to the full-attention portion of its cache.
For the main evaluations, models were run in FP16 on an NVIDIA B200 GPU with frozen base-model weights; only the value-aware indexer was trained.

\paragraph{Indexer training.}
For each model we trained the indexers on 600 context-only articles (plus 60 for development) from 11 public long-context datasets, with a 10-epoch KL-only warm-up and five joint epochs of Equation~\ref{eq:training_objective} with $(\lambda_{\mathrm{out}},\lambda_{\mathrm{KL}})=(2,1)$.
Supervision was generated through repeat-prefill without QA labels, and the trained indexers were frozen for evaluation.
The indexers add 0.57--2.13\% of the backbone parameters (Table~\ref{tab:indexer_size}) and train once per model in 34--151 B200 GPU-hours (Appendix~\ref{app:training_details}).

\paragraph{Benchmarks and metrics.}
We evaluated long-context performance on QuALITY, RULER, and LongBench v1.
For QuALITY~\citep{pangQuALITYQuestionAnswering2022}, we used 50 articles and all 894 associated questions, and reported multiple-choice accuracy.
For RULER~\citep{hsiehRULERRealContext2024}, all methods were evaluated on the same 550 examples at a 4K-token context length using the aggregate official string-match score.
For LongBench v1~\citep{baiLongBenchBilingual2024}, we evaluated all 21 tasks using 50 test instances per task (1,050 instances in total) and the official task-specific metrics.

\paragraph{Baselines and cache budget.}
We compared ARC-KV with SnapKV~\citep{liSnapKVLLMKnows2024}, Expected Attention (EA)~\citep{devotoExpectedAttentionKV2025}, KVzip~\citep{kimKVzipQueryAgnosticKV2025}, and the default OMP-based Attention Matching (AM)~\citep{zweigerFastKVCompaction2026}, plus full-context and no-context references (Appendix~\ref{app:baselines}).
With retention ratio $\rho=t/T$, compacted methods were evaluated at $\rho\in\{0.20,0.10,0.05,0.02,0.01\}$; ARC-KV ratios are nominal under the per-head allocation (Appendix~\ref{sec:method_inference}).
For the Llama main evaluation, we used $\lambda_m=0.25$ on QuALITY and RULER and $0.75$ on LongBench v1.

\subsection{Main Results}

\begin{figure}[t]
\centering
\includegraphics[width=\textwidth]{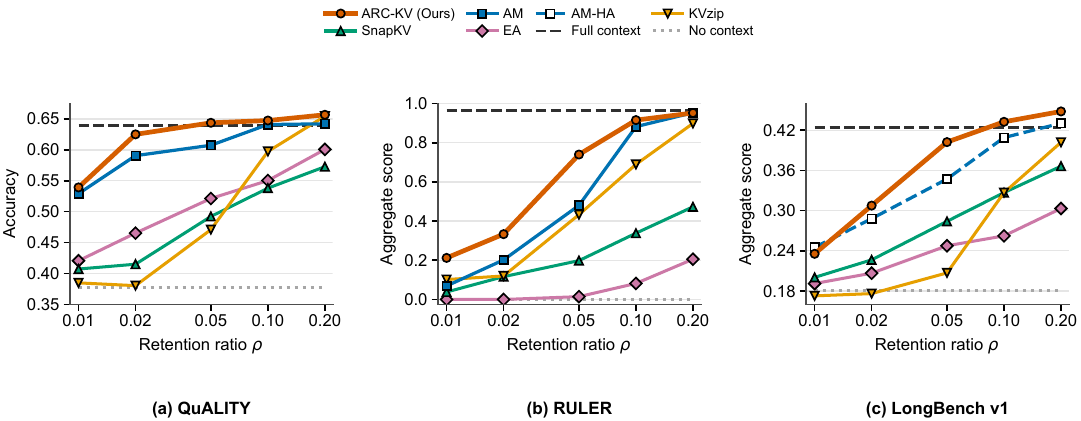}
\caption{Llama-3.1-8B-Instruct results (higher is better): \textbf{(a)} QuALITY accuracy on 894 questions from 50 articles, \textbf{(b)} RULER aggregate string-match score on 550 examples at a 4K-token context length, and \textbf{(c)} LongBench v1 aggregate score over 21 tasks with 50 instances each.
AM uses OMP in \textbf{(a,b)} and highest-attention selection (AM-HA) in \textbf{(c)}.}
\label{fig:main_results}
\end{figure}

ARC-KV ranked first among the reported compacted methods in 13 of the 15 benchmark--retention settings (Figure~\ref{fig:main_results}): all five on QuALITY and four of five on both RULER and LongBench.
The two exceptions were RULER at $\rho=0.20$, where ARC-KV essentially matched AM (0.9515 versus 0.9527), and LongBench at $\rho=0.01$, where it trailed AM-HA (0.2355 versus 0.2451).

ARC-KV improved over the strongest compacted baseline at each retention ratio by 0.23--3.63 percentage points on QuALITY, by 3.34--25.94 points on RULER from $\rho=0.01$ to $0.10$ (e.g., 0.7394 versus 0.4800 for AM at $\rho=0.05$), and by 1.76--5.51 points on LongBench from $\rho=0.02$ to $0.20$.
Relative to full context, ARC-KV reached 0.6437--0.6567 on QuALITY from $\rho=0.05$ to $0.20$ versus 0.6387, retained 98.6\% of the RULER score at $\rho=0.20$, and scored 0.4326 and 0.4481 on LongBench at $\rho=0.10$ and $0.20$ versus 0.4243.
Because the OMP-based AM variant was impractically slow over all 1,050 LongBench instances, we used its highest-attention variant, AM-HA, for this benchmark; Appendix~\ref{app:main_results_details} gives the per-benchmark discussion.

\subsection{Cross-Model Results}

We further evaluated ARC-KV with Qwen3-8B and Gemma-3-12B-IT at a fixed retention ratio of $\rho=0.05$.
As shown in Table~\ref{tab:cross_model_results}, ARC-KV ranked first among the compacted methods in three of the four benchmark--model settings.
On QuALITY, ARC-KV achieved 0.5034 with Qwen3-8B and surpassed AM by 1.01 percentage points.
It achieved 0.6846 with Gemma-3-12B-IT and surpassed the strongest baseline AM by 0.34 points.
On RULER with Gemma-3-12B-IT, ARC-KV reached 0.3954 and outperformed KVzip by 16.81 points.
With Qwen3-8B, ARC-KV instead ranked second on RULER with 0.3815, trailing KVzip (0.4040) by 2.25 points.
Thus, ARC-KV remained competitive across both evaluated architectures, although it did not dominate every benchmark--model setting.

\begin{table}[t]
\caption{Cross-model results at $\rho=0.05$ (higher is better).
AM denotes OMP-based Attention Matching; bold marks the best compacted method in each row.}
\label{tab:cross_model_results}
\centering
\footnotesize
\setlength{\tabcolsep}{4.0pt}
\begin{tabular}{@{}l|c|c|c|c|c|c|c@{}}
\toprule
& \multicolumn{2}{c|}{References} & \multicolumn{5}{c}{Compacted methods} \\
\cmidrule(lr){2-3}\cmidrule(lr){4-8}
Benchmark@Model & Full ctx. & No ctx. & AM & EA & SnapKV & KVzip & \textbf{ARC-KV (Ours)} \\
\midrule
QuALITY@Qwen3-8B       & 0.5537 & 0.2383 & 0.4933 & 0.2830 & 0.2562 & 0.2125 & \textbf{0.5034} \\
RULER@Qwen3-8B         & 0.9764 & 0.0000 & 0.2527 & 0.0109 & 0.0818 & \textbf{0.4040} & 0.3815 \\
QuALITY@Gemma-3-12B-IT & 0.7114 & 0.4944 & 0.6812 & 0.5962 & 0.5403 & 0.4832 & \textbf{0.6846} \\
RULER@Gemma-3-12B-IT   & 0.9727 & 0.0000 & 0.1982 & 0.0509 & 0.1018 & 0.2273 & \textbf{0.3954} \\
\bottomrule
\end{tabular}
\end{table}

\subsection{Efficiency and Ablations}

\begin{figure}[!t]
\centering
\includegraphics[width=0.85\linewidth]{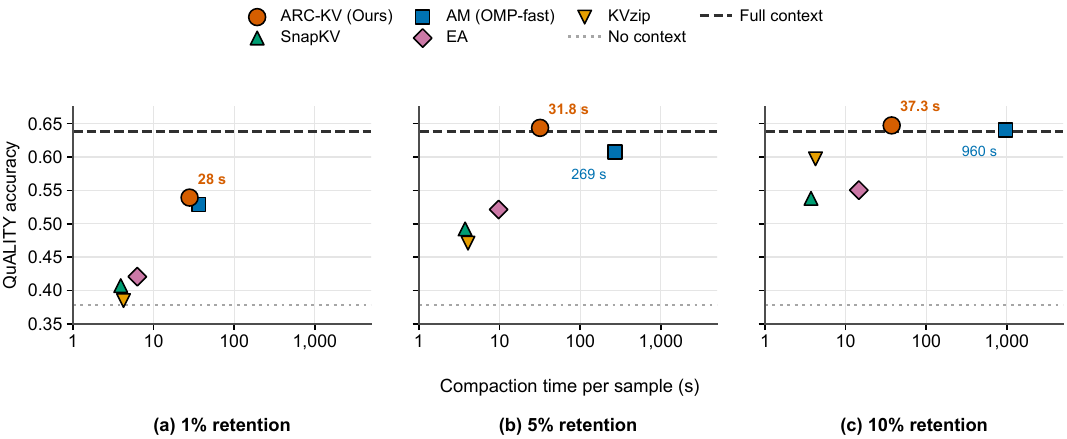}
\caption{QuALITY accuracy versus compaction time for Llama-3.1-8B-Instruct at \textbf{(a)} 1\%, \textbf{(b)} 5\%, and \textbf{(c)} 10\% retention.
The horizontal axis reports per-sample compaction time in seconds on a logarithmic scale.
Times include reported query-generation costs and ARC-KV's reference-query construction; horizontal lines are untimed full- and no-context references.}
\label{fig:quality_efficiency_tradeoff}
\end{figure}

Figure~\ref{fig:quality_efficiency_tradeoff} compares QuALITY accuracy with the reported per-sample cache-compaction time at three representative retention ratios.
ARC-KV was faster and more accurate than reconstruction-based AM at every evaluated budget.
At 1\% retention, ARC-KV reduced compaction time from 36.34 to 28.0~s and improved accuracy by 1.01 percentage points.
At 5\% retention, it reduced time from 269.46 to 31.8~s and improved accuracy by 3.63 points.
At 10\% retention, it reduced time from 959.83 to 37.3~s and improved accuracy by 0.65 points.
These reductions corresponded to speedups of 1.30$\times$, 8.47$\times$ and 25.73$\times$.
The lightweight selectors SnapKV, KVzip and EA remained faster in absolute compaction time.
However, the strongest of these alternatives at each budget was 5.01--12.24 points less accurate than ARC-KV.
These measurements cover cache compaction only and do not represent end-to-end inference latency.

\paragraph{Compaction time breakdown.}
A whole-model profile of all 256 layer--KV-head pairs on synthetic 4K--64K contexts at $\rho=0.05$ (Appendix~\ref{app:time_breakdown}, Figure~\ref{fig:compaction_time_breakdown}) shows OMP key selection growing from 14.932~s to 29.49~min while the learned indexer needs 0.516--7.273~s (28.9--243.3$\times$); totals are 2.35~s and 1.66~min versus 16.47~s and 30.91~min for AM, and at 64K repeat prefill is 70.8~s of ARC-KV's 99.8~s.
At long context lengths, reference-query construction therefore becomes the dominant remaining cost.

\paragraph{Decode-time attention.}
At $\rho=0.20$, compact attention with the fitted bias was 3.19--4.43$\times$ faster per layer than full-cache FlashAttention-2 at 4K--32K contexts while storing about $5\times$ fewer K/V states (Appendix~\ref{app:decode_speedup}, Table~\ref{tab:decode_attention_speedup}); this isolates attention, not end-to-end decode latency.

\paragraph{Ablations.}
On QuALITY with Llama-3.1-8B-Instruct (details in Appendix~\ref{app:ablations}), root-mean-square aggregation beats mean and max pooling by 4.59--8.95 points at $\rho\leq0.05$, and replacing the per-head budget with a uniform one costs 1.68--16.22 points.
Every nonzero $\lambda_m$ lowers the attention-mass error of unmerged anchors (by 43.9--56.9\% at the best setting, with a broad optimum; Table~\ref{tab:key_merging_coefficient_ablation}).

On the first 30 QuALITY articles, the Llama-3.1-8B-Instruct indexer and OMP-fast show limited anchor-set overlap, with mean Jaccard scores of 0.144--0.201 across 1--20\% retention under matched per-head budgets (Appendix~\ref{app:anchor_overlap}).

\section{Related Work}

Token-subset eviction, whether training-free~\citep{xiaoStreamingLLM2024,zhangH2OHeavyHitterOracle2023,liSnapKVLLMKnows2024,kimKVzipQueryAgnosticKV2025,devotoExpectedAttentionKV2025} or learned~\citep{zengIncontextKVCacheEviction2025,moschellaLearningEvictKeyValue2026,zhouLKVEndtoEndLearning2026}, deploys a hard subset $(K_S,V_S)$ without value refitting or attention-mass correction, and compensated-retention methods~\citep{chariKVDistillNearlyLossless2025,yangIndexMemLearnedKVCache2026} add model components instead.
Attention-Gate learns head-specific gates, KVP learns eviction rankings from future-attention rewards, and LKV jointly learns head budgets and token masks through self-distillation.
KV-Distill adapts context encoding so retained states aggregate preceding information, whereas IndexMem stores evicted information in an online latent memory.
ARC-KV leaves the base context encoding unchanged and fits $\beta$ and $C_v$ for each context, placing the compensation in the compact cache.
Appendix~\ref{app:related_work} discusses both lines in detail.

\paragraph{Cache merging, reconstruction, and synthesis.}
Cartridges~\citep{eyubogluCartridgesLightweightGeneralpurpose2025} learns a new compact KV cache for each context, which is expressive but requires costly gradient-based optimization per context.
Attention Matching~\citep{zweigerFastKVCompaction2026} restricts $C_k$ to original keys and fits attention-mass weights and $C_v$ in closed form, but its OMP search with repeated NNLS refitting dominates the cost.
Still~\citep{oneillStillAmortizedKV2026} trains a per-layer Perceiver to synthesize a compact cache in one forward pass, an end-to-end compactor that must be trained separately for each base-model checkpoint.
ARC-KV instead learns only one scalar score per key, which keeps offline training lightweight, and constructs $C_k$, $\beta$, and $C_v$ for each context (Appendix~\ref{app:related_work}).
ARC-KV occupies a distinct point in this design space: it amortizes discrete anchor discovery, while retaining context-specific reconstruction of keys, attention mass, and values.



\section{Conclusion}
We presented ARC-KV, which uses a value-aware indexer to amortize anchor selection while retaining context-specific convex-hull-constrained key merging and attention-mass and compact-value fitting.
On Llama-3.1-8B-Instruct, ARC-KV led reported compacted methods in most settings and led all three benchmark aggregates at 2\% retention.
On QuALITY at 10\% retention, it improved accuracy from 0.6409 to 0.6474 over OMP-based Attention Matching while reducing reported compaction time from 959.8 to 37.3~s.
ARC-KV currently targets one-shot compaction of static reusable context prefixes.
Extending it to generated-state recompression and reducing reference-query construction and dense key-assignment costs remain future work.

\clearpage
\subsection*{AI use statement}
We used large language model tools to polish and organize the manuscript, identify and collect related literature, edit code, and revise figures and charts.
The authors take full responsibility for the content of this paper, including all AI-assisted text, claims, and references.

\bibliography{iclr2027_conference}
\bibliographystyle{iclr2027_conference}

\appendix

\clearpage
\section{Additional Method Details}
\label{app:method_details}

\subsection{Value-Aware Indexer Architecture}
\label{app:indexer_architecture}

For physical KV head $h$, let $G=H_Q/H_{KV}$ and $\mathcal{G}_h=\{hG,\ldots,(h+1)G-1\}$ denote its associated query heads under grouped-query attention.
Each scoring row $i=(r,g)$ pairs a reference position $r$ with a query head $g\in\mathcal{G}_h$, uses its post-RoPE query $q_i\in\mathbb{R}^{d}$, and shares the frozen-model pre-query-projection activation $x_r\in\mathbb{R}^{d_x}$ across the $G$ query heads at position $r$.
The corresponding post-RoPE keys are $k_j\in\mathbb{R}^{d}$, whereas values $v_j\in\mathbb{R}^{d}$ do not receive positional rotation.
For every layer--KV-head pair, the indexer maintains an independent parameter set with $H_I$ index heads of width $d_I$ and one auxiliary value-attention head of width $d_A$:
\begin{equation}
L_q,L_k\in\mathbb{R}^{H_Id_I\times d},
\qquad
L_x\in\mathbb{R}^{H_I\times d_x},
\qquad
b_x\in\mathbb{R}^{H_I}.
\label{eq:indexer_parameters}
\end{equation}
The value-aware pathway adds
\begin{equation}
U_q,U_k,U_v\in\mathbb{R}^{d_A\times d},
\qquad
L_c\in\mathbb{R}^{H_Id_I\times d_A},
\qquad
L_v\in\mathbb{R}^{H_Id_I\times d}.
\label{eq:value_indexer_parameters}
\end{equation}
We first obtain a query-conditioned summary of the value cache:
\begin{equation}
A^{V}_{i,j}
=\frac{\exp\!\left((U_qq_i)^\top(U_kk_j)\right)}
{\sum_{u=1}^{T}\exp\!\left((U_qq_i)^\top(U_kk_u)\right)},
\qquad
c_i=\sum_{j=1}^{T}A^{V}_{i,j}U_vv_j.
\label{eq:value_context}
\end{equation}
This is a single auxiliary attention head shared by all $H_I$ index heads, rather than a separate value-attention operation for each index head.
The value context enriches the query embedding, while each cached value enriches its colocated key embedding:
\begin{equation}
\bar q_{i,a}=\mathcal{N}\!\left((L_q q_i+L_cc_i)_a\right),
\qquad
\bar k_{j,a}=\mathcal{N}\!\left((L_k k_j+L_vv_j)_a\right),
\qquad
w_{i,a}=(L_xx_r+b_x)_a.
\label{eq:indexer_projection}
\end{equation}
Here $a\in[H_I]$, each projected block has width $d_I$, and $\mathcal{N}(z)=z/\sqrt{\lVert z\rVert_2^2/d_I+10^{-6}}$ is parameter-free RMS normalization.
For reference query $i$ and post-RoPE cached key $j$, the indexer logit is
\begin{equation}
I_{i,j}=\sum_{a=1}^{H_I}w_{i,a}\,
\operatorname{LeakyReLU}_{0.1}\!\left(
\frac{\bar q_{i,a}^{\top}\bar k_{j,a}}{\sqrt{d_I}}
\right).
\label{eq:pairwise_index_score}
\end{equation}
The projections $L_c$ and $L_v$ are initialized to zero, whereas $U_q$, $U_k$, and $U_v$ are initialized with entrywise standard deviation $d^{-1/2}$.
Consequently, loading a query--key-only checkpoint initially reproduces its logits exactly, after which optimization can learn value-dependent corrections.
The frozen language-model weights are not modified; only the independent indexer parameters for each layer--KV-head pair are learned.

\subsection{Inference}
\label{sec:method_inference}

At inference, ARC-KV constructs one reusable compact cache for each context.
We first prefill the original context to obtain its full cache $(K,V)$ and length $T$.
We then obtain $\mathcal{D}_{\mathrm{ref}}$ through the repeat-prefill procedure in Section~\ref{sec:method_overview}.
Unlike training, inference does not partition this pool into scoring, fitting, and holdout subsets.
The available pool is reused for anchor scoring, key merging, and attention mass and compact value fitting.

Following prior nonuniform allocation schemes~\citep{liuScissorhands2023,fengAdaKV2025}, we first distribute the nominal cache budget across the compacted layer--KV-head pairs using fixed model-specific weights.
Let $\mathcal{P}$ denote this set of pairs, let $a_{\ell,h}$ be normalized such that $\sum_{(\ell,h)\in\mathcal{P}}a_{\ell,h}=1$, and let $t$ be the nominal per-pair budget.
We assign
\begin{equation}
t_{\ell,h}
=\operatorname{clip}_{[1,T]}\!\left(
\left\lfloor a_{\ell,h}|\mathcal{P}|t\right\rfloor
\right).
\label{eq:per_head_budget}
\end{equation}
Integer rounding and clipping can make the realized aggregate budget differ slightly from its nominal value.

For each layer--KV-head pair, the trained value-aware indexer evaluates the reference pairs and pools its query-specific scores into one score per cache position.
The top-$t_{\ell,h}$ positions form the real-key anchor set $S$.
For long contexts, we instead select anchors within contiguous cache chunks and take their union as $S$, as detailed in Appendix~\ref{app:kv_chunking}.
ARC-KV assigns every cached key to an anchor by similarity between their normalized attention-response profiles and merges each group to form $C_k$.
It then fits the attention-mass bias $\beta$ and compact values $C_v$.
The resulting tuple $(C_k,\beta,C_v)$ defines the compact cache in Equation~\ref{eq:compact_attention} and is reused for future queries to the same context.

The value-aware indexer therefore replaces iterative OMP anchor search for each layer--KV-head pair with one scoring pass.
Appendix~\ref{app:complexity} reports the time and memory complexity of these operations.
Our current scope is one-shot compaction of a reusable context prefix.

\subsection{KV-Based Chunking for Long Contexts}
\label{app:kv_chunking}

Let $T_a$ denote the length of the article span to be compacted, excluding any prefix or suffix states retained verbatim.
When fixed chunking is enabled and $T_a>10{,}000$, we partition the score vector into contiguous chunks of configured length $L_{\mathrm{chunk}}$ and retain the local top $t_{\ell,h,c}=\operatorname{clip}_{[1,T_c]}(\operatorname{round}(t_{\ell,h}T_c/T_a))$ anchors from each chunk.
The union of these anchors defines $S$; response-profile assignment, convex key merging, and the fits of $\beta$ and $C_v$ are then performed once against the full article cache rather than independently within each chunk.
This rounding rule can make the realized anchor count differ slightly from $t_{\ell,h}$ and from the ceiling rule used by the output-training surrogate, while fixed chunking is bypassed when $T_a\leq10{,}000$.

\subsection{Complexity Analysis}
\label{app:complexity}

For $n$ scoring queries, the indexer evaluates the auxiliary value attention and low-dimensional index scores in $O\!\left(nT(H_Id_I+2d_A)\right)$ time, followed by one top-budget anchor selection.
The merge uses all $n$ queries; constructing response profiles and mass weights, assigning all keys to $t$ anchors, and accumulating the centroids require $O\!\left(nT(d+t)+Td\right)$ time and $O(nT+Tt)$ intermediate memory in the current dense implementation.
The learned scorer removes the repeated support selection and coefficient refitting performed across the $t$ rounds of OMP, but reference-query construction, dense key-to-anchor assignment, and the final mass and value solves remain, with assignment potentially dominating at larger budgets.

\subsection{Method Overview and Reference-Query Construction}
\label{app:reference_queries}

\paragraph{Method summary.}
ARC-KV replaces iterative, per-context anchor search with a value-aware indexer that selects real-key anchors in one scoring pass.
The indexer is trained once per base model using supervision derived from full-attention outputs, after which its frozen parameters are reused across contexts.
For each new context, ARC-KV applies the trained indexer, merges discarded keys into the selected anchors, and analytically fits an attention-mass bias and compact values.
Throughout this section, we describe one layer--KV-head pair under grouped-query attention (GQA). The same procedure is applied independently to all layer--KV-head pairs.

\paragraph{Learned and context-specific components.}
Only the indexer parameters are learned.
Section~\ref{sec:indexer_training} describes how full-attention supervision trains the indexer to select anchors that preserve held-out attention outputs after analytical fitting.
Once trained, the same frozen indexer is reused across contexts, while reference-query construction, scoring, merging, and fitting are performed separately for each context.
Figure~\ref{fig:method_overview} summarizes the training and cache-construction procedures.

\paragraph{Reference-query pool.}
Reference queries should probe the cache after the model has observed the complete context.
We obtain such queries through the repeat-prefill procedure used by KVzip~\citep{kimKVzipQueryAgnosticKV2025} and Attention Matching~\citep{zweigerFastKVCompaction2026}.
An initial prefill produces the full cache $(K,V)$.
We then append the instruction ``Repeat the previous context verbatim.'' using the model's chat template and teacher-force the original context as the assistant response.
At each repeated-context position, we retain the pre-query-projection activation $x_i$ and its post-RoPE query $q_i$.
These pairs form $\mathcal{D}_{\mathrm{ref}}=\{(x_i,q_i)\}_{i=1}^{n_{\mathrm{ref}}}$, with $X_{\mathrm{ref}}$ and $Q_{\mathrm{ref}}$ denoting the matrices that row-stack the activations and queries, respectively.
The repeated copy supplies only the reference pairs, while all keys and values being compacted come from the original-context prefill.
Because these queries revisit the context after the model has observed it in full, they provide context-conditioned probes of the original cache.

\paragraph{Full caption of Figure~\ref{fig:method_overview}.}
Overview of ARC-KV training and inference.
During offline training, full-attention behavior supervises a reusable value-aware indexer.
A hard top-$t$ branch evaluates the selected anchors, while a differentiable soft surrogate supplies gradients to the indexer.
At inference, the frozen indexer scores a new cache once to select real-key anchors.
ARC-KV then merges the corresponding key groups and fits the attention-mass bias $\beta$ and compact values $C_v$ to materialize a reusable compact cache.
The indexer parameters are shared across contexts, whereas all cache-construction operations are context-specific.

\subsection{Value-Aware Indexer: Design Walk-Through}
\label{app:indexer_walkthrough}

To make anchor selection sensitive to value-side information, we instantiate the indexer as a value-aware extension of the low-dimensional query--key scorer used by the DeepSeek Sparse Attention indexer~\citep{deepseek-aiDeepSeekV32PushingFrontier2025}.
A single auxiliary value-attention head forms a query-conditioned summary of the value cache, which augments the query-side features.
The indexer represents each cache position using both its key $k_j$ and value $v_j$.
For each reference query $i$, it derives weights from $x_i$ to combine the scores produced by its low-dimensional index heads.
This design allows two keys with similar attention patterns to receive different scores when they carry different value-side information.
The complete architecture and initialization are given in Appendix~\ref{app:indexer_architecture}.
Section~\ref{sec:indexer_training} describes the reconstruction-based procedure used to train and reuse this scorer.

\paragraph{Anchor selection.}
We select the $t$ positions with the largest aggregate scores as real-key anchors:
\begin{equation}
S=\operatorname*{TopK}_{j\in[T]}(s_j,t).
\label{eq:persistent_support}
\end{equation}
We write $K_S\equiv K_{S,:}$ for the selected real keys.
These anchors define the subsequent key groups but need not equal the final compact keys $C_k$.
Unlike query-conditioned sparse attention, ARC-KV selects this set once per context and layer--KV-head pair and reuses it for all future queries.

\subsection{Joint Preservation Objective}
\label{app:joint_objective}

\paragraph{Full-cache attention.}
For queries $Q\in\mathbb{R}^{n\times d}$, the full-cache attention probabilities and outputs are
\begin{equation}
P(Q;K)=\operatorname{softmax}\!\left(QK^\top\right),
\qquad
O(Q;K,V)=P(Q;K)V.
\label{eq:full_attention}
\end{equation}

\paragraph{Joint objective with an uncompressed block.}
For an uncompressed KV block $(K_{\mathrm{fixed}},V_{\mathrm{fixed}})$ of length $F$, such as retained prompt states or subsequently generated states, let $b=[\beta;\mathbf{0}_F]$ append zero bias to its logits.
The joint preservation objective can then be written as
\begin{equation}
\frac{\exp\!\left(q\begin{bmatrix}K\\K_{\mathrm{fixed}}\end{bmatrix}^{\!\top}\right)
\begin{bmatrix}V\\V_{\mathrm{fixed}}\end{bmatrix}}
{\sum_{j=1}^{T+F}\exp\!\left(q\begin{bmatrix}K\\K_{\mathrm{fixed}}\end{bmatrix}_{j}^{\!\top}\right)}
\;\approx\;
\frac{\exp\!\left(q\begin{bmatrix}C_k\\K_{\mathrm{fixed}}\end{bmatrix}^{\!\top}+b^\top\right)
\begin{bmatrix}C_v\\V_{\mathrm{fixed}}\end{bmatrix}}
{\sum_{j=1}^{t+F}\exp\!\left(q\begin{bmatrix}C_k\\K_{\mathrm{fixed}}\end{bmatrix}_{j}^{\!\top}+b_j\right)}.
\label{eq:concatenated_attention_preservation}
\end{equation}

\subsection{Key-Merging Schematic and Remarks}
\label{app:key_merging_details}

\begin{figure}[H]
\centering
\includegraphics[width=0.75\linewidth]{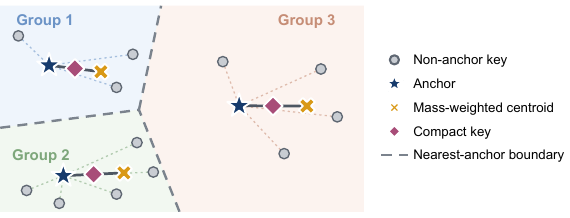}
\caption{Schematic of convex-hull-constrained key merging.
Non-anchor keys are grouped by normalized attention-response similarity, and query-averaged unnormalized attention mass defines the group centroid $\mu_r$.
The compact key $c_{k,r}$ interpolates between $\mu_r$ and the selected real-key anchor ($\lambda_m=0.5$ shown).
The two-dimensional regions schematically represent assignment in attention-response-profile space rather than Euclidean distance between the plotted keys.}
\label{fig:key_merging_schematic}
\end{figure}

Figure~\ref{fig:key_merging_schematic} summarizes the grouping and merging procedure.

\paragraph{Tie-breaking.}
Ties for non-anchor keys are resolved by a fixed deterministic rule.

\paragraph{Key-merging coefficient.}
Here, $\lambda_m$ is the key-merging coefficient that controls how far each selected anchor moves toward its attention-mass-weighted group centroid.
The endpoint $\lambda_m=0$ leaves the anchor keys unmerged, so $C_k=K_S$, while $\lambda_m=1$ replaces each anchor by its group centroid.

\subsection{Attention Mass and Compact Value Fitting}
\label{sec:cache_reconstruction}

Given the merged $C_k$ and a calibration-query matrix $Q_{\mathrm{fit}}$ drawn from $\mathcal{D}_{\mathrm{ref}}$, we apply the data-dependent calibration introduced by Attention Matching~\citep{zweigerFastKVCompaction2026}.
Let $n_{\mathrm{fit}}$ be the number of rows in $Q_{\mathrm{fit}}$, let $z_i^{\mathrm{fit}}=\max_j(q_i^\top k_j)$, and let $z^{\mathrm{fit}}=[z_i^{\mathrm{fit}}]_i$.
We define $\Phi_K=\exp(Q_{\mathrm{fit}}K^\top-z^{\mathrm{fit}}\mathbf{1}_T^\top)$ and $\Phi_C=\exp(Q_{\mathrm{fit}}C_k^\top-z^{\mathrm{fit}}\mathbf{1}_t^\top)$.
We first approximate the nonnegative attention-mass fit with an unconstrained least-squares solve followed by elementwise projection:
\begin{equation}
\begin{aligned}
\widetilde w
&=\operatorname*{arg\,min}_{w\in\mathbb{R}^{t}}
\left\|\Phi_Cw-\Phi_K\mathbf{1}_T\right\|_2^2,\\
w^\star&=\max\{\widetilde w,\mathbf{0}_t\},\\
\beta^\star
&=\operatorname{clip}_{[b_{\min},b_{\max}]}\!\left(
\log\!\left(\max\{w^\star,\epsilon_w\mathbf{1}_t\}\right)
\right).
\end{aligned}
\label{eq:mass_fit}
\end{equation}
Here, both maxima are elementwise, $\epsilon_w>0$ prevents taking the logarithm of zero, and $[b_{\min},b_{\max}]$ provides a finite numerical guard.
With $\widehat P=\operatorname{softmax}(Q_{\mathrm{fit}}C_k^\top+\mathbf{1}_{n_{\mathrm{fit}}}(\beta^\star)^\top)$ and $Y=\operatorname{softmax}(Q_{\mathrm{fit}}K^\top)V$, we then fit
\begin{equation}
C_v^\star=\operatorname*{arg\,min}_{C\in\mathbb{R}^{t\times d}}
\left\|\widehat PC-Y\right\|_F^2+\lambda_v\lVert C\rVert_F^2.
\label{eq:value_lstsq}
\end{equation}
Here, $\lambda_v\geq0$ is the compact-value ridge coefficient.
Together, the two fits approximate the original cache's unnormalized attention mass and attention outputs without adding learned parameters.

\subsection{Indexer Training: Additional Derivations}
\label{app:indexer_training_details}

\paragraph{Training overview.}
The indexer architecture in Section~\ref{sec:key_indexer} is deliberately lightweight.
The central learning contribution is the training formulation that turns this scorer into a reusable anchor-selection policy.
We train one indexer per frozen base model using the model's full-attention behavior as self-supervision, without anchor labels generated by iterative search.
Training begins with a KL warm-up that teaches query-specific token relevance.
Joint training then optimizes reconstruction of held-out attention outputs through a hard-selection forward branch and a differentiable soft surrogate for backpropagation.
Both branches operate on real keys before key merging, so the learned objective targets anchor quality directly.
Deterministic grouping and key merging are applied afterward when constructing the deployed compact cache.

\paragraph{Held-out attention targets.}
For a query $q$, define the full-cache attention distribution and output as
\begin{equation}
p(q)=\operatorname{softmax}\!\left(qK^\top\right),
\qquad
y(q)=p(q)V.
\label{eq:holdout_attention}
\end{equation}

\paragraph{Soft surrogate branch.}
Because top-$t$ selection is non-differentiable, the backward branch relaxes the decision over the candidate set $\mathcal{C}=\operatorname{TopK}(s,\min\{T,2t\})$, which contains $S$.
Let $\operatorname{sg}$ denote stop-gradient and let $s_{(t)}$ be the hard-selection threshold.
For each $j\in\mathcal{C}$, we define
\begin{equation}
\tau=\operatorname{sg}(s_{(t)}),
\qquad
\sigma_s=\operatorname{sg}(\operatorname{std}(s)),
\qquad
g_j=\sigma\!\left(\frac{s_j-\tau}{\gamma_e\sigma_s}\right),
\label{eq:soft_candidate_gate}
\end{equation}
where $\sigma(\cdot)$ is the logistic sigmoid.
The gate softly relaxes membership near the top-$t$ boundary and is not normalized to enforce $\sum_jg_j=t$.
During backpropagation, we treat $\mathcal{C}$, $\tau$, and $\sigma_s$ as constants, so gradients reach the indexer through the candidate scores in $g$.
The temperature $\gamma_e$ is annealed linearly from $1.0$ to $0.1$ over joint training.

The soft branch uses the real candidate keys $C_k^{\mathrm{s}}=K_{\mathcal{C},:}$.
For an anchor candidate $j=s_r\in S$, we set its base bias to $b_j^{\mathrm{s}}=\beta_r^{\mathrm{h}}$.
Every additional candidate receives $b_j^{\mathrm{s}}=0$.
The effective bias is $b^{\mathrm{s}}+\log(g)$, which gives
\begin{equation}
\widehat p^{\mathrm{s}}(q)
=\operatorname{softmax}\!\left(
q(C_k^{\mathrm{s}})^\top+(b^{\mathrm{s}}+\log(g))^\top
\right).
\label{eq:soft_attention_output}
\end{equation}
Using the corresponding attention matrix on $Q_{\mathrm{fit}}$, we obtain $C_v^{\mathrm{s}}$ with a differentiable ridge solve and set $\widehat y^{\mathrm{s}}(q)=\widehat p^{\mathrm{s}}(q)C_v^{\mathrm{s}}$.
We evaluate $\mathcal{L}_{\mathrm{out}}^{\mathrm{s}}$ on $Q_{\mathrm{holdout}}$ using the same squared relative-error form as Equation~\ref{eq:output_loss}.

\paragraph{KL warm-up and joint objective.}
Before joint training, we warm up the indexer by matching its query-specific key distribution to full attention on $\mathcal{D}_{\mathrm{ref}}$:
\begin{equation}
\mathcal{L}_{\mathrm{KL}}
=\frac{1}{|\mathcal{D}_{\mathrm{ref}}|}
\sum_{(x,q)\in\mathcal{D}_{\mathrm{ref}}}
\operatorname{KL}\!\left(
p(q)\,\middle\|\,\operatorname{softmax}(I_\theta(x,q,K,V))
\right).
\label{eq:indexer_kl}
\end{equation}

\section{Additional Motivation Details}
\label{app:motivation_details}

\subsection{Anchor Selector Definitions}
\label{app:value_selector_details}

The motivating selector comparison uses disjoint query sets for anchor scoring and held-out evaluation.
For $n$ scoring queries, let $p_{ij}=\operatorname{softmax}_j(q_iK^\top)$ and $O_i=\sum_{u=1}^{T}p_{iu}v_u$.
Highest-attention ranking pools the attention probabilities, whereas value-influence ranking pools each position's normalized leave-one-out output change:
\begin{equation}
\begin{aligned}
D_{ij}
&=\left\|O_i-O_i^{(-j)}\right\|_2
=\frac{p_{ij}}{1-p_{ij}}\left\|v_j-O_i\right\|_2,\\
s_j^{\mathrm{HA}}
&=\left(\frac{1}{n}\sum_i p_{ij}^2\right)^{1/2},
&
s_j^{\mathrm{VI}}
&=\left(\frac{1}{n}\sum_i
\frac{D_{ij}^2}{\left\|O_i\right\|_2^2+\epsilon}
\right)^{1/2}.
\end{aligned}
\label{eq:background_value_selectors}
\end{equation}
Here, $O_i^{(-j)}$ is the attention output after removing position $j$ and renormalizing over the remaining positions.
The top-budget positions under $s^{\mathrm{HA}}$ or $s^{\mathrm{VI}}$ define the corresponding anchor set.

\paragraph{Evaluation protocol (from Section~\ref{sec:background}).}
Traditional KV cache compaction methods rank keys from attention probabilities alone, ignoring value information~\citep{zhangH2OHeavyHitterOracle2023,liSnapKVLLMKnows2024}. We represent this family with Highest-Attention ranking (HA), which selects keys with the largest aggregated attention scores, without considering their value vectors. A separate line of work incorporates value-side information when estimating token importance~\citep{guoVATP2024,goelCAOTE2025}; we represent this family with Value-Influence ranking (VI), which selects keys by their normalized leave-one-out effect on the attention output and therefore explicitly accounts for values (Appendix~\ref{app:value_selector_details}).

To test whether values help select keys, we rank the same cache with HA and VI under a matched budget, then evaluate the selected sets on disjoint held-out queries.
For a selected set $S$, we measure the full-cache output contribution it fails to capture:
\begin{equation}
\widetilde O_i(S)=\sum_{j\in S}p_{ij}v_j,
\qquad
E(S)=\frac{1}{|Q_{\mathrm{eval}}|}\sum_i
\frac{\left\|O_i-\widetilde O_i(S)\right\|_2}
{\left\|O_i\right\|_2+\epsilon}.
\label{eq:background_truncated_error}
\end{equation}
Here, $p_{ij}$ and $O_i$ are computed from the full cache, so lower $E(S)$ indicates better key selection.
Across the four selected layer--KV-head pairs for visual clarity in Figure~\ref{fig:motivation_results}a, VI reduces held-out error at every keep ratio, by 1.05--6.23\% relative to HA.
This consistent gain shows that values provide useful information beyond attention alone when selecting keys.

\subsection{Hard-Subset and Reconstructed Attention Operators}
\label{app:background_operators}

For simplicity, all attention equations omit the standard inverse-square-root scaling of query--key logits.

To compare the fidelity of cache reconstruction under token eviction methods versus cache reconstruction, we track the attention output of each and compare it against the full-cache attention. 
Token eviction methods (rule-based or learned) construct a compact cache by selecting an index set $S$ and retaining the corresponding KV pairs $(K_S,V_S)$ unchanged, where $K_S=[k_j]_{j\in S}$ and $V_S=[v_j]_{j\in S}$. For a query $q\in\mathbb{R}^{d}$, these retained states produce
\begin{equation}
\widehat O_{\mathrm{hard}}(q;S)
=\operatorname{softmax}\!\left(qK_S^\top\right)V_S
\label{eq:background_hard_attention}
\end{equation}
Reconstruction-based methods, instead use the selected keys as anchors to construct compact keys $C_k$, an additive mass bias $\beta$, and compact values $C_v$ from the full cache.
With $C_k,C_v\in\mathbb{R}^{|S|\times d}$ and $\beta\in\mathbb{R}^{|S|}$, the same query is evaluated as
\begin{equation}
\widehat O_{\mathrm{rec}}(q;C_k,\beta,C_v)
=\operatorname{softmax}\!\left(qC_k^\top+\beta^\top\right)C_v
\label{eq:background_reconstructed_attention}
\end{equation}
Each row of $C_k$ corresponds to one selected anchor, $\beta$ calibrates its attention mass, and the corresponding row of $C_v$ supplies its compact value.

\subsection{Attention Matching Pipeline Profile}
\label{app:am_profile}
To identify the computational bottleneck of the AM method~\citep{zweigerFastKVCompaction2026}, we profile the complete AM pipeline using the authors' official open-source implementation. AM compacts a cache in four stages. It first constructs a reference-query set $Q_{\mathrm{ref}}$ from the input context.
It then applies OMP~\citep{patiOrthogonalMatchingPursuit1993} to select an anchor set $S$ from the original keys.
With $K_S$ fixed, AM performs a final fit of $\beta$ to match the full cache's attention mass and then fits $C_v$ to reconstruct its attention outputs. 
Accordingly, we break compaction time into four parts: reference-query construction (context prefill), iterative anchor search, $\beta$ fitting, and $C_v$ fitting.

\begin{figure}[H]
\centering
\includegraphics[width=\textwidth]{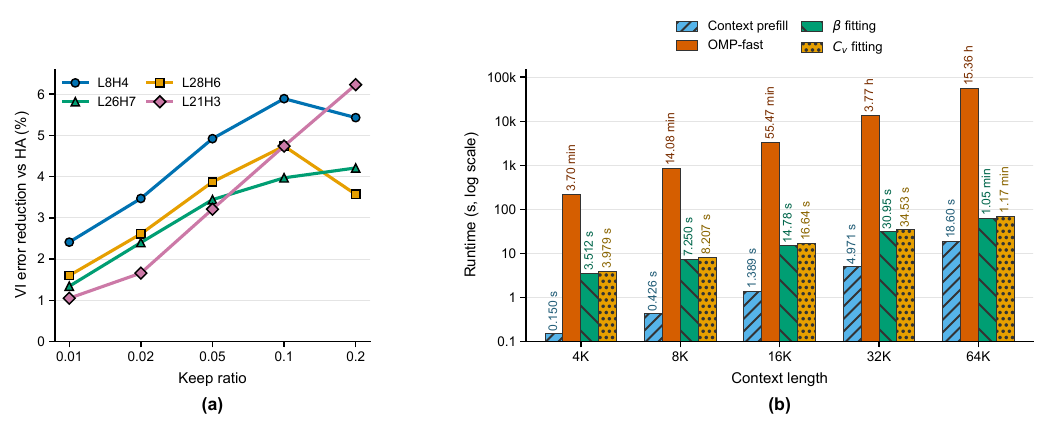}
\caption{
\textbf{(a)} Percentage reduction in held-out truncated-output relative error achieved by VI over HA for four selected layer--KV-head pairs on Llama-3.1-8B-Instruct and QuALITY.
At keep ratio $\rho$, for layer $\ell$ and KV head $h$, the plotted quantity is $\bigl(E_{\ell,h}(S_{\mathrm{HA}}(\rho))-E_{\ell,h}(S_{\mathrm{VI}}(\rho))\bigr)/E_{\ell,h}(S_{\mathrm{HA}}(\rho))$.
Positive values favor VI.
L$x$H$y$ denotes layer $x$ and KV head $y$.
\textbf{(b)} Runtime of context prefill, OMP-fast anchor search, nonnegative mass-weight fitting, and compact-value fitting across context lengths at a 20\% retention ratio on Llama-3.1-8B-Instruct.
Runtime is shown on a logarithmic scale, and every bar is annotated with its measured value.}
\label{fig:motivation_results}
\end{figure}

As shown in Figure~\ref{fig:motivation_results}, iterative anchor search dominates the profile and scales much more steeply than the other stages.
The OMP-fast search alone requires 3.70 minutes at 4K tokens and 15.36 hours at 64K tokens. At 64K, each remaining profiled stage takes at most 1.17 minutes.
This gap arises from dependent selection and mass-refitting steps.
The principal opportunity for amortization therefore lies in anchor search, not in the final context-specific reconstruction.

\subsection{Motivation Figure}
\label{app:motivation_figure}

\section{Additional Experimental Setup}
\label{app:setup_details}

\subsection{Model and Implementation Details}
\label{app:model_details}

Because Gemma-3-12B-IT uses hybrid attention, we compacted only its eight full-attention layers.
Its 40 sliding-window layers were left unchanged because each caches at most 1,024 tokens.
The eight full-attention layers therefore account for 44\% of the KV cache at 4K tokens and 93\% at 64K.
All models were run in FP16 on an NVIDIA B200 GPU.
The base-model weights remained frozen for ARC-KV, and only the value-aware indexer was trained.
At inference time, ARC-KV used the repeat-prefill reference-query construction described in Section~\ref{sec:method_overview} to build a reusable compact cache for each context.

\subsection{Indexer Training Details}
\label{app:training_details}

\paragraph{Indexer training.}
Our training sources comprised 11 datasets spanning multihop reasoning (HotpotQA, MuSiQue, and 2WikiMultihopQA)~\citep{yangHotpotQADataset2018,trivediMuSiQueMultihopQuestions2022,hoConstructingMultihopQADataset2020}, summarization (GovReport, Multi-News, and QMSum)~\citep{huangEfficientAttentionsLong2021,fabbriMultiNewsLargeScale2019,zhongQMSumNewBenchmark2021}, code (RepoBench-Python v1.1 and LCC-Python)~\citep{liuRepoBenchBenchmarkingRepositoryLevel2024,guoLongCoderLongRange2023}, and question answering (TriviaQA, NarrativeQA, and Qasper)~\citep{joshiTriviaQALargeScale2017,kociskyNarrativeQAReadingComprehension2018,dasigiDatasetInformationSeekingQuestions2021}.
From the source training splits, we randomly downsampled each category to 150 training articles and 15 development articles, yielding 600 training and 60 development articles in total.
We chose this scale to keep indexer training within about 150 GPU-hours per model.
Under the Qwen3-8B tokenizer, these 660 articles contained approximately 4.24M tokens in total, with individual contexts ranging from 1{,}102 to 16{,}884 tokens.
We retained only the context text and generated supervision through repeat-prefill without QA labels.
We used the same two-stage schedule for all three models.
For the first 10 epochs, we optimized Equation~\ref{eq:training_objective} with $(\lambda_{\mathrm{out}},\lambda_{\mathrm{KL}})=(0,1)$ as a KL-only warm-up.
We then trained for five additional epochs with $(\lambda_{\mathrm{out}},\lambda_{\mathrm{KL}})=(2,1)$, corresponding to $2\mathcal{L}_{\mathrm{out}}^{\mathrm{STE}}+\mathcal{L}_{\mathrm{KL}}$.
Both stages used AdamW with $(\beta_1,\beta_2)=(0.9,0.999)$, weight decay 0.01, and global gradient-norm clipping at 1.0.
The peak learning rates were $10^{-4}$ for the KL-only stage and $3\times10^{-5}$ for the joint stage.
In each stage, the learning rate increased linearly over the first 3\% of steps and then followed cosine decay to 10\% of its peak value.

\begin{table}[H]
\caption{Total trainable parameter count of the value-aware indexers across all compacted layer--KV-head pairs, with $H_I=8$ and $d_I=d_A=128$ per pair. The Pairs column reports the number of independently parameterized layer--KV-head indexers, and ratios are relative to each model's text backbone. Gemma includes only the eight full-attention layers used for compaction.}
\label{tab:indexer_size}
\centering
\footnotesize
\setlength{\tabcolsep}{3.5pt}
\begin{tabular}{@{}l|r|r|r|r@{}}
\toprule
Model & Pairs & Total indexer & Base model & Ratio \\
\midrule
Llama-3.1-8B     & 256 & 155M & 8.03B  & 1.93\% \\
Qwen3-8B         & 288 & 175M & 8.19B  & 2.13\% \\
Gemma-3-12B-IT   &  64 &  67M & 11.77B & 0.57\% \\
\bottomrule
\end{tabular}
\end{table}

\paragraph{Training cost.}
Training was layer-sharded across eight independent NVIDIA B200 GPU jobs (183~GB each), without inter-GPU communication.
Including initial reference-query construction, total costs were 147 GPU-hours for Llama, 151 GPU-hours for Qwen, and 34 GPU-hours for Gemma.
Peak per-GPU memory was 35.4, 40.3, and 23.7~GB, respectively.
Each indexer was trained once per base model and then frozen for reuse across inference contexts.

\subsection{Baselines and Cache Budget}
\label{app:baselines}

\paragraph{Baselines.}
We compared ARC-KV with two reference conditions and five KV-cache compaction baselines.
\emph{Full context} retained the complete context cache without compaction, whereas \emph{No context} omitted the document context and evaluated the model from the task input alone.
SnapKV ranked prompt states using attention from a prompt-final observation window~\citep{liSnapKVLLMKnows2024}.
Attention Matching (AM) used its default OMP-based variant, which greedily selected original keys to reconstruct the full-cache attention mass and then fitted attention-mass weights and compact values~\citep{zweigerFastKVCompaction2026}.
KV Policy (KVP) learned per-head eviction rankings from future-attention rewards~\citep{moschellaLearningEvictKeyValue2026}.
KVzip scored cached states through teacher-forced context reconstruction~\citep{kimKVzipQueryAgnosticKV2025}, whereas Expected Attention estimated their contributions under a model of the future-query distribution~\citep{devotoExpectedAttentionKV2025}.

\paragraph{Cache budget.}
For a layer--KV-head cache with original length $T$ and target compacted length $t$, we defined the retention (keep) ratio as $\rho=t/T$.
We evaluated each compacted method at $\rho\in\{0.20,0.10,0.05,0.02,0.01\}$.
For ARC-KV, we distributed the nominal budget across layer--head pairs using the fixed model-specific allocation described in Section~\ref{sec:method_inference}; reported retention ratios are nominal values.
Full context corresponded to $\rho=1$ and was never compacted. Its score is therefore independent of the compacted-cache budget and is repeated across retention-ratio columns only as a common reference.

\section{Additional Results}
\label{app:additional_results}

\subsection{Per-Benchmark Discussion of the Main Results}
\label{app:main_results_details}

\paragraph{QuALITY.}
As shown in Figure~\ref{fig:main_results}(a), ARC-KV improved over the strongest compacted baseline at each retention ratio by 0.23--3.63 percentage points.
The gains were most pronounced at moderate-to-low budgets: ARC-KV exceeded AM by 3.63 points at $\rho=0.05$ and by 3.44 points at $\rho=0.02$.
Moreover, from $\rho=0.05$ to $0.20$, ARC-KV achieved accuracies of 0.6437--0.6567, exceeding the full-context reference of 0.6387 while retaining only 5--20\% of the cache.
Thus, under these moderate retention budgets, compaction did not impose an accuracy penalty and slightly improved the point estimates over full-context inference.

\paragraph{RULER.}
On RULER (Figure~\ref{fig:main_results}(b)), ARC-KV outperformed all compacted baselines from $\rho=0.01$ to $0.10$, with margins of 3.34--25.94 percentage points over the strongest competitor at each budget.
At $\rho=0.05$, for example, ARC-KV reached 0.7394 compared with 0.4800 for AM.
At $\rho=0.20$, ARC-KV approached the full-context score, attaining 0.9515 versus 0.9655 and thereby retaining 98.6\% of full-context performance after removing 80\% of the cache.
On RULER, ARC-KV therefore combined near-full-context performance at $\rho=0.20$ with substantial gains over compacted baselines at lower retention ratios.

\paragraph{LongBench.}
Because the OMP-based AM variant was impractically slow over all 1,050 LongBench instances, we used its highest-attention variant, AM-HA, for this benchmark.
As shown in Figure~\ref{fig:main_results}(c), ARC-KV achieved the highest aggregate score from $\rho=0.02$ to $0.20$, exceeding the strongest reported compacted baseline by 1.76--5.51 percentage points.
At $\rho=0.01$, ARC-KV scored 0.2355 versus 0.2451 for AM-HA.
ARC-KV also exceeded the full-context point estimate at $\rho=0.10$ and $0.20$ (0.4326 and 0.4481 versus 0.4243), while outperforming SnapKV, EA, and KVzip at every evaluated ratio.

\paragraph{Full caption of Figure~\ref{fig:main_results}.}
Main results for Llama-3.1-8B-Instruct across three long-context benchmarks and five KV-cache retention ratios (higher is better).
\textbf{(a)}, QuALITY multiple-choice accuracy on 894 questions from 50 articles; ARC-KV ranked first among reported compacted methods at every ratio.
\textbf{(b)}, RULER aggregate string-match score on 550 examples at a 4K-token context length; ARC-KV ranked first at four of five ratios and trailed AM only at $\rho=0.20$.
\textbf{(c)}, LongBench v1 aggregate score over 21 tasks with 50 test instances per task; ARC-KV ranked first at four of five ratios and trailed AM-HA only at $\rho=0.01$.
EA denotes Expected Attention. AM denotes the default OMP-based Attention Matching variant in \textbf{(a)} and \textbf{(b)}, whereas AM-HA denotes its highest-attention variant in \textbf{(c)}.
ARC-KV uses $\lambda_m=0.25$ in \textbf{(a)} and \textbf{(b)} and $\lambda_m=0.75$ in \textbf{(c)}; the full- and no-context lines are budget-independent references.

\subsection{Quality--Efficiency Trade-off: Additional Remarks}
\label{app:tradeoff_details}

Compared with OMP-based AM, ARC-KV substantially reduced compaction time.
It also achieved higher accuracy than the faster lightweight selectors.
However, these measurements cover cache compaction only and do not represent end-to-end inference latency.

\paragraph{Full caption of Figure~\ref{fig:quality_efficiency_tradeoff}.}
Accuracy--time trade-off for Llama-3.1-8B-Instruct on QuALITY at \textbf{(a)} $\rho=0.01$, \textbf{(b)} $\rho=0.05$, and \textbf{(c)} $\rho=0.10$.
Each point represents a compacted-cache method; the horizontal lines show the full- and no-context accuracy references, which do not perform cache compaction and are therefore not treated as timed scatter points.
The horizontal axis uses a logarithmic scale and reports per-sample compaction time in seconds.
The baseline totals already include their reported query-generation time, whereas ARC-KV uses the single-deployment total including reference-query construction.

\subsection{Compaction Time Breakdown}
\label{app:time_breakdown}

\begin{figure}[H]
\centering
\includegraphics[width=\textwidth]{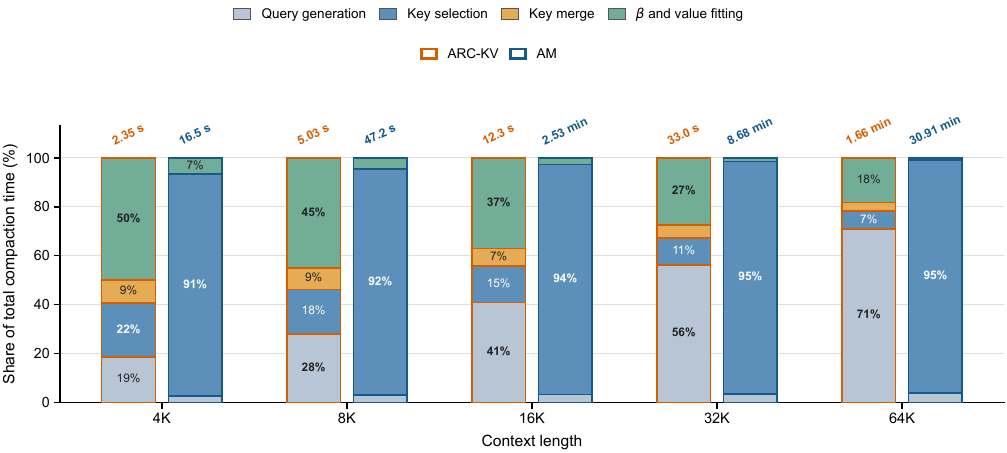}
\caption{Whole-model cache-compaction time for Llama-3.1-8B-Instruct on synthetic contexts at $\rho=0.05$.
Timings cover all 256 layer--KV-head pairs and were measured in FP16 on an NVIDIA B200 GPU.
Each context length contains one stacked bar for ARC-KV and one for AM; fill colors identify stages and outline colors identify methods.
Each bar is normalized to 100\%, so segment heights show the percentage of total compaction time attributable to each stage; labels above the bars report the corresponding absolute totals.
Query generation uses repeat prefill for both methods.
For ARC-KV, the remaining segments are learned-indexer top-$t$ selection, mass-weighted key merging, and the sum of guarded attention-mass fitting and merged-key least squares.
For AM, they are OMP-fast selection and the sum of the reported baseline NNLS and unmerged least-squares fitting times.
OMP-fast uses $k=4$ and refits every two selection steps.
The plotted totals are sums of the measured components rather than independent end-to-end measurements.}
\label{fig:compaction_time_breakdown}
\end{figure}

Figure~\ref{fig:compaction_time_breakdown} profiles whole-model cache construction for all 256 layer--KV-head pairs of Llama-3.1-8B-Instruct on synthetic 4K--64K contexts at $\rho=0.05$.
Each context length contains paired stacked bars for ARC-KV and AM, with repeat prefill treated as the query-generation stage for both methods.
OMP key selection increased from 14.932~s at 4K to 29.49~min at 64K, whereas the learned indexer required only 0.516--7.273~s over the same range.
Replacing iterative search therefore reduced selection time by 28.9$\times$ at 4K and 243.3$\times$ at 64K.

The component-wise total for ARC-KV was 2.35~s at 4K and 1.66~min at 64K, compared with 16.47~s and 30.91~min for AM, corresponding to 7.01$\times$ and 18.59$\times$ speedups.
At 64K, repeat prefill accounted for 70.8~s of ARC-KV's 99.8~s total, whereas learned key selection required only 7.273~s.
Thus, amortizing anchor selection removes the iterative-search bottleneck and shifts the dominant remaining cost to reference-query generation at long context lengths.

\subsection{Decode-Time Attention Speedup}
\label{app:decode_speedup}

KV-cache compaction reduces both memory use and the cost of attending to the cached context during decoding.
We measured one decode step with Llama-3.1-8B-Instruct at batch size 64 on an NVIDIA B200 GPU in BF16.
The full-cache baseline used FlashAttention-2, whereas ARC-KV retained 20\% of the KV states and applied its fitted attention-mass bias $\beta$.
Measurements used eager execution and were reproducible within approximately 1--3\%.
Table~\ref{tab:decode_attention_speedup} reports the per-layer latency and the corresponding sum over all 32 attention layers.

\begin{table}[H]
\caption{Decode-time attention efficiency with a full cache and an ARC-KV cache at $\rho=0.20$.
K/V memory and latency are reported per attention layer for a batch of 64.
The latency column compares full-cache FlashAttention-2 with compact attention including the fitted bias $\beta$.
The final column sums the per-layer latency over all 32 attention layers.}
\label{tab:decode_attention_speedup}
\centering
\scriptsize
\setlength{\tabcolsep}{2.5pt}
\begin{tabular}{@{}l|c|c|c|c@{}}
\toprule
Context & \shortstack{K/V per layer (GB)\\Full $\rightarrow$ compact}
& \shortstack{Latency per layer ($\mu$s)\\FA2 $\rightarrow$ ARC-KV}
& Speedup
& \shortstack{32-layer attention\\(ms/token)} \\
\midrule
4K  & $1.07 \rightarrow 0.21$ & $174.3 \rightarrow 54.6$  & $3.19\times$ & $5.58 \rightarrow 1.75$ \\
8K  & $2.15 \rightarrow 0.43$ & $341.4 \rightarrow 89.4$  & $3.82\times$ & $10.92 \rightarrow 2.86$ \\
16K & $4.29 \rightarrow 0.86$ & $667.9 \rightarrow 166.3$ & $4.02\times$ & $21.37 \rightarrow 5.32$ \\
32K & $8.59 \rightarrow 1.72$ & $1472.5 \rightarrow 332.2$ & $4.43\times$ & $47.12 \rightarrow 10.63$ \\
\bottomrule
\end{tabular}
\end{table}

At $\rho=0.20$, ARC-KV reduced the stored K/V states by approximately $5\times$.
Per-layer attention latency improved by $3.19\times$ at 4K and $4.43\times$ at 32K.
Across 32 layers, the attention component decreased from 5.58 to 1.75~ms per token at 4K and from 47.12 to 10.63~ms at 32K.
The full-length bias-aware path was only $1.06$--$1.09\times$ slower than FlashAttention-2.
The measured speedup therefore came primarily from shortening the attended cache rather than omitting the fitted bias.
These measurements isolate attention and do not represent end-to-end decode latency.
CUDA Graph capture may change the absolute latency but does not change the cache-size reduction.

\subsection{Anchor-Selection Overlap with OMP}
\label{app:anchor_overlap}

We compared the trained Llama-3.1-8B-Instruct indexer with OMP-fast on the first 30 articles of the QuALITY development split, covering all 32 layers and eight KV heads at retention ratios $\rho\in\{0.01,0.02,0.05,0.10,0.20\}$.
This diagnostic uses the same anchor count $k=\lfloor\rho T_a\rfloor$ for both selectors in each layer--KV-head pair, where $T_a$ is the article length; budgets are uniform across pairs, unlike the default nonuniform allocation in Section~\ref{sec:method_inference}.
The articles contain 2,722--8,094 tokens and each occupies one chunk under the configured chunk size of 8,192.
Anchor positions are measured relative to the article, excluding prompt and suffix tokens.

The run used repeat-prefill reference queries with a query budget of 2,048, a query-construction budget of 50,000, and seed 0.
The recorded KV states came from BF16 model execution and were cast to FP32 for this diagnostic.
The indexer used softmax normalization followed by RMS aggregation.
The recorded OMP-fast settings were \texttt{k\_choice=4}, \texttt{nnls\_interval=2}, and \texttt{nnls\_iters=0}.

\begin{figure}[t]
\centering
\includegraphics[width=\textwidth]{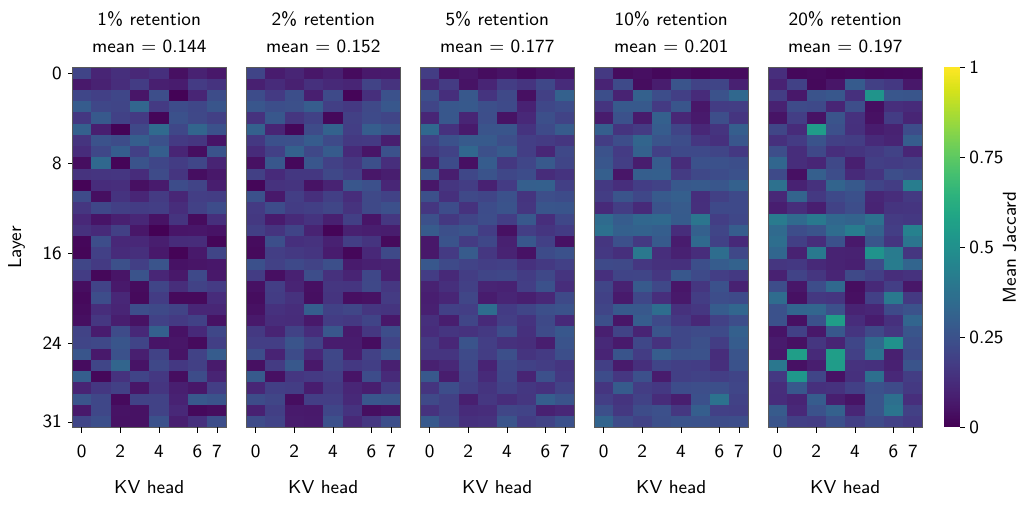}
\caption{Anchor-set overlap between the Llama-3.1-8B-Instruct indexer and OMP-fast on the first 30 QuALITY development articles under matched uniform per-head budgets.
Each cell is the mean Jaccard overlap over the 30 articles for one layer--KV-head pair; layer and head indices are zero-based.
The five retention ratios share a color scale from 0 (disjoint sets) to 1 (identical sets).
Panel means average all 7,680 article--layer--KV-head observations at that ratio.
The comparison concerns selected real-key anchors before key merging and the final cache-reconstruction fits.}
\label{fig:anchor_overlap}
\end{figure}

For each article--layer--KV-head observation, let $S_A$ and $S_O$ be the indexer's and OMP-fast's selected token-index sets before key merging and the final cache-reconstruction fits, with $|S_A|=|S_O|=k$.
We measure Jaccard overlap and the fraction of selected anchors shared by the two methods as
\begin{equation}
J=\frac{|S_A\cap S_O|}{|S_A\cup S_O|},
\qquad
R=\frac{|S_A\cap S_O|}{k}.
\label{eq:anchor_overlap}
\end{equation}
Figure~\ref{fig:anchor_overlap} averages $J$ over the 30 articles separately for each layer--KV-head pair.
Overall means give equal weight to each of the $30\times32\times8=7{,}680$ observations per retention ratio; $R$ is likewise computed per observation before averaging.

The mean Jaccard scores are 0.144, 0.152, 0.177, 0.201, and 0.197 at the five increasing retention ratios.
The mean shared-anchor fraction $R$ ranges from 24.1\% to 32.6\%, indicating that the two selectors often choose different supports at matched cardinality.
Overlap varies across layers and heads: at 20\% retention, the article-averaged Jaccard scores range from 0.012 to 0.558 across the 256 pairs.
This diagnostic characterizes agreement between anchor identities; it does not establish a downstream quality advantage for either selector or isolate the causes of ARC-KV's task-level results.

\section{Ablation Studies}
\label{app:ablations}

\subsection{Key-Merging Coefficient}

We varied the key-merging coefficient $\lambda_m$ in Equation~\ref{eq:key_merge} while holding the remaining ARC-KV components fixed.
For each layer--KV-head pair, let $Z_f=\Phi_K\mathbf{1}_T$ and $Z_c=\Phi_Cw^\star$ denote the full- and compact-cache unnormalized attention-mass responses in Equation~\ref{eq:mass_fit}.
We measured the relative attention-mass error
$E_{\mathrm{mass}}=\lVert Z_c-Z_f\rVert_2/\lVert Z_f\rVert_2$ on the 50 QuALITY articles with Llama-3.1-8B-Instruct.
Because the full-response magnitudes vary by orders of magnitude across layer--KV-head pairs, we normalize each error by $\lVert Z_f\rVert_2$ before aggregation rather than report raw $\ell_2$ error.

\begin{table}[H]
\caption{Relative attention-mass $\ell_2$ error for different key-merging coefficients $\lambda_m$ on 50 QuALITY articles with Llama-3.1-8B-Instruct (lower is better).
Bold indicates the minimum at each retention ratio at the reported precision.}
\label{tab:key_merging_coefficient_ablation}
\centering
\small
\setlength{\tabcolsep}{6pt}
\begin{tabular}{@{}l|c|c|c|c|c@{}}
\toprule
$\lambda_m$ & $\rho=0.20$ & $\rho=0.10$ & $\rho=0.05$ & $\rho=0.02$ & $\rho=0.01$ \\
\midrule
0.0 & 0.791 & 1.019 & 1.279 & 1.505 & 1.539 \\
\midrule
0.2 & 0.534 & 0.625 & 0.745 & 0.887 & 0.948 \\
0.3 & 0.460 & 0.523 & 0.632 & 0.792 & 0.879 \\
0.4 & 0.417 & 0.470 & 0.584 & \textbf{0.764} & \textbf{0.864} \\
0.5 & 0.395 & 0.446 & \textbf{0.571} & \textbf{0.764} & 0.868 \\
0.6 & 0.386 & \textbf{0.439} & 0.573 & 0.774 & 0.877 \\
0.7 & \textbf{0.384} & 0.440 & 0.582 & 0.786 & 0.886 \\
1.0 & 0.392 & 0.456 & 0.613 & 0.821 & 0.911 \\
\bottomrule
\end{tabular}
\end{table}

Every nonzero $\lambda_m$ reduced the relative attention-mass error compared with the unmerged anchors ($\lambda_m=0$) at every retention ratio.
The best setting reduced the error by 43.9--56.9\%, with the minimizing coefficient increasing from $0.4$ at $\rho=0.01$ to $0.7$ at $\rho=0.20$.
At $\rho=0.02$, $\lambda_m=0.4$ and $0.5$ were tied at the reported precision.
At every ratio, at least one neighboring setting was within 0.004 of the minimum, indicating a broad optimum rather than sensitivity to one exact coefficient.
These results show that moving anchors toward their group centroids consistently improves attention-mass reconstruction, while the preferred coefficient increases with the retention ratio.

\subsection{Aggregation Method}

We compared the root-mean-square aggregation in Equation~\ref{eq:indexer_pooling} against mean and max aggregation.
It applies a softmax over keys for each reference query before pooling each key across queries.
Table~\ref{tab:aggregation_ablation} reports the change in QuALITY accuracy relative to root-mean-square aggregation.
Negative values indicate lower accuracy.

\begin{table}[H]
\caption{Aggregation ablation on QuALITY with Llama-3.1-8B-Instruct.
Entries are accuracy differences in percentage points relative to root-mean-square aggregation across KV-cache retention ratios (higher is better).
An asterisk marks a statistically significant difference from root-mean-square aggregation.}
\label{tab:aggregation_ablation}
\centering
\small
\setlength{\tabcolsep}{5pt}
\begin{tabular}{@{}l|c|c|c|c|c@{}}
\toprule
Aggregation & 0.20 & 0.10 & 0.05 & 0.02 & 0.01 \\
\midrule
Root-mean-square & 0.00 & 0.00 & 0.00 & 0.00 & 0.00 \\
Mean & $-1.12$ & $-3.69^{*}$ & $-6.60^{*}$ & $-5.37^{*}$ & $-4.59^{*}$ \\
Max  & $-1.45$ & $-2.24$ & $-7.27^{*}$ & $-8.95^{*}$ & $-7.05^{*}$ \\
\bottomrule
\end{tabular}
\end{table}

Root-mean-square aggregation achieved the highest accuracy at every retention ratio.
At $\rho\leq0.05$, mean aggregation was worse by 4.59--6.60 points and max aggregation by 7.05--8.95 points.
This pattern is consistent with max aggregation being sensitive to a single noisy query-specific peak.
Root-mean-square aggregation first normalizes scores across keys for each query.
It then pools the bounded responses across queries.
This preserves strong evidence from a subset of queries without allowing one maximum response to determine the ranking.

\subsection{Per-head Budget Allocation}

We ablated the fixed model-specific allocation across layer--KV-head pairs described in Section~\ref{sec:method_inference} by replacing it with a uniform allocation.
Both variants used the same total nominal cache budget and differed only in how that budget was distributed across layer--head pairs.
Table~\ref{tab:per_head_budget_ablation} reports the change in QuALITY accuracy relative to the default per-head allocation.

\begin{table}[H]
\caption{Per-head budget-allocation ablation on QuALITY with Llama-3.1-8B-Instruct.
Entries are accuracy differences in percentage points relative to the default model-specific per-head allocation across KV-cache retention ratios (higher is better).
Both variants use the same total nominal cache budget.
An asterisk marks a statistically significant difference from the default allocation.}
\label{tab:per_head_budget_ablation}
\centering
\small
\setlength{\tabcolsep}{5pt}
\begin{tabular}{@{}l|c|c|c|c|c@{}}
\toprule
Allocation & 0.20 & 0.10 & 0.05 & 0.02 & 0.01 \\
\midrule
Per-head (default) & 0.00 & 0.00 & 0.00 & 0.00 & 0.00 \\
Uniform & $-1.68$ & $-4.59^{*}$ & $-10.29^{*}$ & $-16.22^{*}$ & $-10.96^{*}$ \\
\bottomrule
\end{tabular}
\end{table}

Uniform allocation reduced QuALITY accuracy at every retention ratio, with losses ranging from 1.68 to 16.22 percentage points.
The penalty exceeded 10 points at all ratios below 10\% retention and was largest at $\rho=0.02$.

\section{Extended Related Work}
\label{app:related_work}

\paragraph{Token-subset cache eviction.}
Training-free eviction methods bound cache growth by retaining selected original KV pairs according to positional or observed-attention signals.
StreamingLLM~\citep{xiaoStreamingLLM2024} preserves initial attention sinks and a recent window, whereas Scissorhands~\citep{liuScissorhands2023}, H2O~\citep{zhangH2OHeavyHitterOracle2023}, and Keyformer~\citep{adnanKeyformer2024} estimate importance from different forms of historical attention.
SnapKV~\citep{liSnapKVLLMKnows2024} uses attention from a prompt-final observation window to rank historical tokens and retain the highest-scoring KV pairs.
These inexpensive policies can fail when positional, historical, or visible-query proxies do not predict a later attention shift.
Reusable-prefix compression must select a context-level cache before any downstream task query is available.
KVzip~\citep{kimKVzipQueryAgnosticKV2025} scores KV pairs through teacher-forced context reconstruction, Compactor~\citep{chariCompactorCalibratedQueryAgnostic2025} combines approximate leverage with non-causal context-attention scores, and Expected Attention~\citep{devotoExpectedAttentionKV2025} estimates KV pair contributions under a Gaussian model of future queries.
These proxy objectives average future-query utility and can therefore miss rare queries whose required tokens receive low proxy scores.
PyramidKV~\citep{caiPyramidKV2024} assigns a fixed pyramidal budget across layers, while AdaKV~\citep{fengAdaKV2025} reallocates each layer's fixed budget across KV heads.
These refinements improve resource allocation, but most subset methods still bind each selected key to its original value and provide no post-selection correction for discarded softmax mass or value contributions.
Our method improves attention-output fidelity by using selected real keys as anchors, merging each anchor group into a compact key, and fitting $\beta$ and $C_v$ against the full cache rather than retaining the corresponding original KV pairs unchanged.

\paragraph{Learned and compensated retention.}
Learned eviction methods replace hand-designed importance rules with model-specific selectors.
Attention-Gate~\citep{zengIncontextKVCacheEviction2025} learns head-specific gates, KVP~\citep{moschellaLearningEvictKeyValue2026} learns per-head rankings from future-attention rewards, and LKV~\citep{zhouLKVEndtoEndLearning2026} jointly learns head budgets and token masks through self-distillation.
These methods learn which original KV entries to retain, but their deployed caches remain hard subsets $(K[S],V[S])$ without value refitting or attention-mass correction.
Other methods preserve discarded information through additional model components: KV-Distill~\citep{chariKVDistillNearlyLossless2025} adapts context encoding so retained states aggregate preceding information, whereas IndexMem~\citep{yangIndexMemLearnedKVCache2026} stores evicted information in an online latent memory.
Our method instead leaves the base context encoding unchanged and fits $\beta$ and $C_v$ for each context, placing the compensation directly in the reusable compact cache.

\paragraph{Cache merging, reconstruction, and synthesis (full discussion).}
Cartridges~\citep{eyubogluCartridgesLightweightGeneralpurpose2025} learns a new compact KV cache for each context by matching the full-context model on synthetic self-study examples.
Because its cache entries are freely optimized rather than selected from the original cache, it is expressive but requires costly gradient-based optimization for every new context.
Attention Matching~\citep{zweigerFastKVCompaction2026} restricts $C_k$ to original keys, fits nonnegative attention-mass weights by NNLS, and fits $C_v$ by least squares.
But its OMP process improves mass reconstruction through repeated greedy selection and NNLS refitting, making support search the dominant cost.
Still~\citep{oneillStillAmortizedKV2026} trains a per-layer Perceiver to synthesize a compact cache in one forward pass, thereby avoiding per-context optimization.
This one-pass inference efficiency comes at the cost of an end-to-end compactor that requires non-trivial offline training resources and must be trained separately for each base-model checkpoint.
Our method scores the original keys and selects real-key anchors in a single indexer pass, making anchor selection efficient.
It then constructs $C_k$ by merging each anchor group and fits $\beta$ and $C_v$ for each context to preserve fidelity to the full-cache attention output.
Because the indexer learns only one scalar score per key rather than synthesizing the compact cache, it requires lightweight offline training.


\end{document}